\pdfoutput=1

\documentclass{article}

\usepackage{hyperref}

\PassOptionsToPackage{numbers, compress}{natbib}

\usepackage[preprint]{neurips26/neurips_2026}

\makeatletter
\newif\ifinbeamer
\@ifclassloaded{beamer}{\inbeamertrue}{\inbeamerfalse}
\makeatother

\ifinbeamer\else
  \usepackage{hyperref}
\fi

\ifinbeamer\else
  \usepackage{xr-hyper}
\fi

\usepackage{amsmath}
\usepackage{mathtools}
\usepackage{amssymb,mathrsfs} 
\usepackage{amssymb}
\usepackage{amsthm}
\usepackage{amsfonts}
\usepackage{upgreek}
\usepackage{nicefrac}
\usepackage{graphicx}
 \usepackage{color}
\usepackage{stmaryrd}
\DeclareMathAlphabet{\mathpzc}{OT1}{pzc}{m}{it}
\ifinbeamer\else
  \usepackage[inline]{enumitem}
\fi
\usepackage{url}
\usepackage{graphicx} 
\usepackage{tikz}
 \usepackage{pgfplots}  
\usepackage{xcolor}
\usepackage{bbm,bm}
\usetikzlibrary{shapes.geometric, arrows.meta, positioning, quotes, chains, calc, fit,backgrounds}

\usepackage{microtype}
\usepackage{ifthen}
\usepackage{xargs}
\ifinbeamer
  \PassOptionsToPackage{compatibility=false}{caption}
\fi
\usepackage{caption}
\usepackage{subcaption}

\usepackage{wrapfig}
\usepackage{makecell}

\usepackage{multirow}

\usepackage{booktabs}       

\usepackage{amsopn}
\usepackage{aliascnt}
\usepackage[capitalize,noabbrev]{cleveref}
\usepackage{autonum}
\usepackage{version}
\excludeversion{commenta}
\usepackage{dsfont}
\usepackage[colorinlistoftodos,prependcaption,textsize=tiny]{todonotes}
\usepackage[most]{tcolorbox}
\usepackage{listings}
\tcbset{
  genbox/.style={
    enhanced,breakable,
    colback=lightgray!5, colframe=lightgray!35!black, boxrule=0.4pt,
    listing only, listing engine=listings,
    listing options={
      basicstyle=\ttfamily\small,
      breaklines=true, breakatwhitespace=false,
      columns=fullflexible, keepspaces=true
    }
  }
}

\usepackage{xcolor}
\usepackage{eso-pic}

\usepackage[framemethod=TikZ]{mdframed}
\mdfdefinestyle{propFrame}{
    linecolor=black!8!white,
    outerlinewidth=.6pt,
    roundcorner=5pt,
    innertopmargin=7pt,
    innerbottommargin=7pt,
    innerrightmargin=8pt,
    innerleftmargin=8pt,
    backgroundcolor=black!3!white
}

\mdfsetup{
  skipabove=.2cm,
  skipbelow=0pt
}

\ifinbeamer\else
\hypersetup{
  colorlinks = true,
  citecolor = blue,
  linkcolor= blue,
  urlcolor= blue,
}
\fi

\definecolor{codegreen}{rgb}{0,0.6,0}
\definecolor{codegray}{rgb}{0.5,0.5,0.5}
\definecolor{codelightgray}{rgb}{0.96,0.97,0.97}
\definecolor{codepurple}{rgb}{0.58,0,0.82}
\definecolor{backcolour}{rgb}{0.95,0.95,0.92}

\lstdefinestyle{mystyle}{
    backgroundcolor=\color{codelightgray},
    commentstyle=\color{codegreen},
    keywordstyle=\color{magenta},
    numberstyle=\tiny\color{codegray},
    stringstyle=\color{codepurple},
    basicstyle=\ttfamily\footnotesize,
    breakatwhitespace=false,         
    breaklines=true,                 
    captionpos=b,                    
    keepspaces=true,                 
    numbers=left,                    
    numbersep=5pt,                  
    showspaces=false,                
    showstringspaces=false,
    showtabs=false,                  
    tabsize=2
}

\makeatletter
\@ifundefined{theorem}{
  \theoremstyle{plain}
  \newtheorem{theorem}{Theorem}[section]
}{}
\@ifundefined{proposition}{\newtheorem{proposition}[theorem]{Proposition}}{}
\@ifundefined{lemma}{}{}
\@ifundefined{corollary}{}{}

\@ifundefined{definition}{
  \theoremstyle{definition}
  
}{}
\@ifundefined{assumption}{}{}

\@ifundefined{remark}{
  \theoremstyle{remark}
  \newtheorem{remark}[theorem]{Remark}
}{}
\makeatother

\crefname{figure}{figure}{figures}
\Crefname{figure}{Figure}{Figures}

\def\mse{\mathsf{E}}

\def\mE{\mathbb{E}}
\def\rset{\mathbb{R}}

\def\rmd{\mathrm{d}}

\newcommand{\dd}{\mathrm{d}}

\newcommandx{\functionspace}[2][1=+]{\mathbb{F}_{#1}(#2)}

\def\l{\left}
\def\r{\right}

\newcommand{\normal}{\mathcal{N}}

\def\Idd{\mathrm{I}_d}

\def\eqsp{\;}
\def\dd{\rmd}

\newcommand{\processd}[1]{\{#1\}_{t = 0}^{T}}

\def\lambday{\lambda^{\latspace}}
\def\lambdayelbo{{\lambda}^{\latspace, \text{E}}}

\def\lambdax{\lambda^{\sspace}}
\def\lambdaxelbo{{\lambda}^{\sspace, \text{E}}}

\def\loss{\mathrm{L}}
\def\lossy{\loss^{\latspace}}
\def\lossx{\loss^{\sspace}}

\def\lambdalatent{\lambda^{\text{latent}}}

\def\entropy{\mathrm{H}}

\def\1{\mathbbm{1}}

\def\Idd{\mathrm{I}_d}

\def\mask{\mathbf{m}}

\def\maskvec{{m}}

\def\bsigma{\bar \sigma}

\def\q{q}

\def\qy{q^{\latspace}}
\def\qx{q^{\sspace}}
\def\qz{q^{\zspace}}
\def\qtildez{\Tilde{q}}
\def\qtildezi{\Tilde{q}^i}

\def\qxi{q^{\sspace, i}}

\def\ptilde{\Tilde{p}}
\def\ptheta{p^{\theta}}
\def\pthetax{p^{\sspace, \theta}}
\def\pthetaxi{p^{\sspace, \theta, i}}
\def\pthetay{p^{\latspace, \theta}}

\def\normal{\mathrm{N}}
\def\sspace{\mathsf{X}}
\def\latspace{\mathsf{Y}}

\def\zspace{\mathsf{Z}}

\def\enc{\mathcal{E}}

\def\KL{\mathrm{KL}}

\def\siglat{\sigma_{\text{lat}}}

\def\xpred{x_{\theta}}
\def\ypred{y_{\theta}}

\def\betavq{\beta}

\newcommandx{\unsure}[2][1=]{\todo[linecolor=red,backgroundcolor=red!25,bordercolor=red,#1]{#2}}
\newcommandx{\change}[2][1=]{\todo[linecolor=blue,backgroundcolor=blue!25,bordercolor=blue,#1]{#2}}
\newcommandx{\info}[2][1=]{\todo[linecolor=OliveGreen,backgroundcolor=OliveGreen!25,bordercolor=OliveGreen,#1]{#2}}
\newcommandx{\improvement}[2][1=]{\todo[linecolor=Plum,backgroundcolor=Plum!25,bordercolor=Plum,#1]{#2}}
\newcommandx{\thiswillnotshow}[2][1=]{\todo[disable,#1]{#2}}

\newcommandx{\dario}[2][1=]{\todo[linecolor=orange,backgroundcolor=orange!25,bordercolor=orange,#1]{#2}}

\newcommandx{\stefano}[2][1=]{\todo[linecolor=red,backgroundcolor=red!25,bordercolor=red,#1]{#2}}

\newcommandx{\alain}[2][1=]{\todo[linecolor=blue,backgroundcolor=blue!25,bordercolor=blue,#1]{#2}}
\def\simplex{\boldsymbol{\Delta}}

\title{Latent-Augmented Discrete Diffusion Models}

\author{
  Dario Shariatian\\
  Inria, PSL Research University\\
  Paris, France\\
  \texttt{dario.shariatian@inria.fr}
  \And
  Alain Durmus\\
  CMAP, Ecole Polytechnique\\
  Palaiseau, France\\
  \texttt{alain.durmus@polytechnique.edu}
  \And
  Umut Simsekli\\
  Inria, PSL Research University\\
  Paris, France\\
  \texttt{umut.simsekli@inria.fr}
  \And
  Stefano Peluchetti\\
  Sakana AI\\
  Tokyo, Japan\\
  \texttt{stepelu@sakana.ai}
}

\begin{document}

\maketitle

\begin{abstract}
Discrete diffusion models have emerged as a powerful class of models and a promising route to fast language generation, but practical implementations typically rely on factored reverse transitions ignoring cross-token dependencies and degrading few-step performance. We propose Latent-Augmented Discrete Diffusion (LADD), which introduces a learnable auxiliary latent channel and performs diffusion over the joint (token, latent) space. The latent variables provide an intermediate representation expressing joint structure while preserving tractable parameterizations.
We instantiate LADD with continuous latents (Co-LADD) and discrete latents (Di-LADD), and study two inference schedules: a joint diffusion that denoises data and latents together, and a sequential diffusion that first resolves latents and then samples tokens conditionally. We derive ELBO-style objectives and analyze design choices that balance latent expressivity with diffusion compatibility. 
In experiments, LADD models yield improvements on unconditional generation metrics as compared to state-of-the-art masked discrete diffusion baselines, and are effective at lower sampling budgets, where unmasking many tokens per step is desirable.

\end{abstract}

\section{Introduction}

Diffusion models have become a standard tool for high-fidelity generation on continuous modalities such as images and audio \citep{esser2024scalingrectifiedflowtransformers, ho2020denoisingdiffusionprobabilisticmodels, chen2021wavegrad}. For discrete data, diffusion offers coarse-to-fine generation with parallel token refinement, flexible conditioning/infilling via bidirectional attention, and opportunities for faster sampling through updating many tokens per step.

Early approaches adapted continuous diffusion to categorical data by embedding tokens in $\rset^d$ and running a continuous process in the embedding space \citep{dieleman2022continuous}. Subsequent work emphasized staying in discrete state space and reported improved performance and scaling characteristics. Within discrete formulations, masked processes are particularly effective and scalable in high dimensions \citep{sahoo2024simpleeffectivemaskeddiffusion, lou2024discrete, hoogeboom2021argmax, austin2021structured}.

However, discrete data, like text, remains challenging for discrete diffusion. Despite steady progress and large-model training \citep{nie2025largelanguagediffusionmodels, song2025seeddiffusionlargescalediffusion}, auto-regressive (AR) models remain strong baselines across model sizes, while discrete diffusion primarily brings improvements by enabling multi-token unmasking per step, enabling faster sampling speeds \citep{campbell2022continuous, lou2024discrete, hoogeboom2021argmax, austin2021structured, zheng2025masked, ni2025trainingoptimallargediffusion, kang2025parallelbenchunderstandingtradeoffsparallel}.

A key limitation in this context is the use of a factored model: given the noise sequence, the distributions of masked tokens are modeled independently of each other, entirely discarding joint structure. This effect is all the more severe in the few-step regime, where independent per-token decisions can combine into jointly inconsistent outputs \citep{liu2025discretecopuladiffusion}. This leads to an incompressible tradeoff: efficient sampling encourages unmasking many tokens per step, but per-token factorization becomes less reliable as this number grows. Predictor–corrector samplers and tailored unmasking schedules help but do not fully address this sensitivity \citep{wang2025remaskingdiscretediffusionmodels, gat2024discreteflowmatching, liu2025thinkgeneratediscretediffusion, shi2024simplified, pham2025discrete}. In continuous diffusion, denoising is soft and reversible, allowing errors to amortize, whereas unveiling a token in masked discrete diffusion is a hard commitment.

\paragraph{Our approach.} We augment masked discrete diffusion with a separate latent channel, which carries cross-token information and supplies a secondary learning and inference signal. Concretely, we introduce \emph{Latent-Augmented Discrete Diffusion} models (LADD), which couple a masked discrete diffusion over tokens with a diffusion over latent embeddings. Latents can be obtained from a pre-trained encoder, or via a custom encoder trained end to end, and ultimately improve coherence when many tokens are unmasked simultaneously.

We instantiate two core variants of this general latent augmentation framework. We explore \textbf{Co-LADD}, which exploit \textbf{Co}ntinuous latent embeddings, and \textbf{Di-LADD}, which exploit \textbf{Di}screte latent tokens. On top of these core model choices, we explore a joint process strategy, where tokens and latents are jointly co-evolving at each step, and a sequential process strategy, where the model produces clean latents first, and then conditions token denoising on them, akin in spirit to \citet{xu2024discodiff}. We derive ELBO-style objectives 
and discuss practical design choices that make each variant deployable.

\paragraph{Contributions.}  \emph{(i)} We make explicit how factored reverse transitions limit masked discrete diffusion in the few-step regime, and highlight differences from continuous models. \emph{(ii)} We propose \emph{Latent-Augmented Discrete Diffusion} models (LADD), latent-augmented processes that leverage joint token structure while retaining parallelism. \emph{(iii)} We derive ELBO-style training objectives and outline design choices for stable optimization. \emph{(iv)} We show that LADD shifts the compute–quality frontier as compared to masked diffusion; in the few step regime, Co-LADD-64 can reduce sampling compute up to $\times 1.5-2.0$ at iso-quality on OWT over a matched-size MDLM \citep{sahoo2024simpleeffectivemaskeddiffusion}.

\begin{figure}[ht!]
    \centering
    \includegraphics[width=0.75\linewidth]{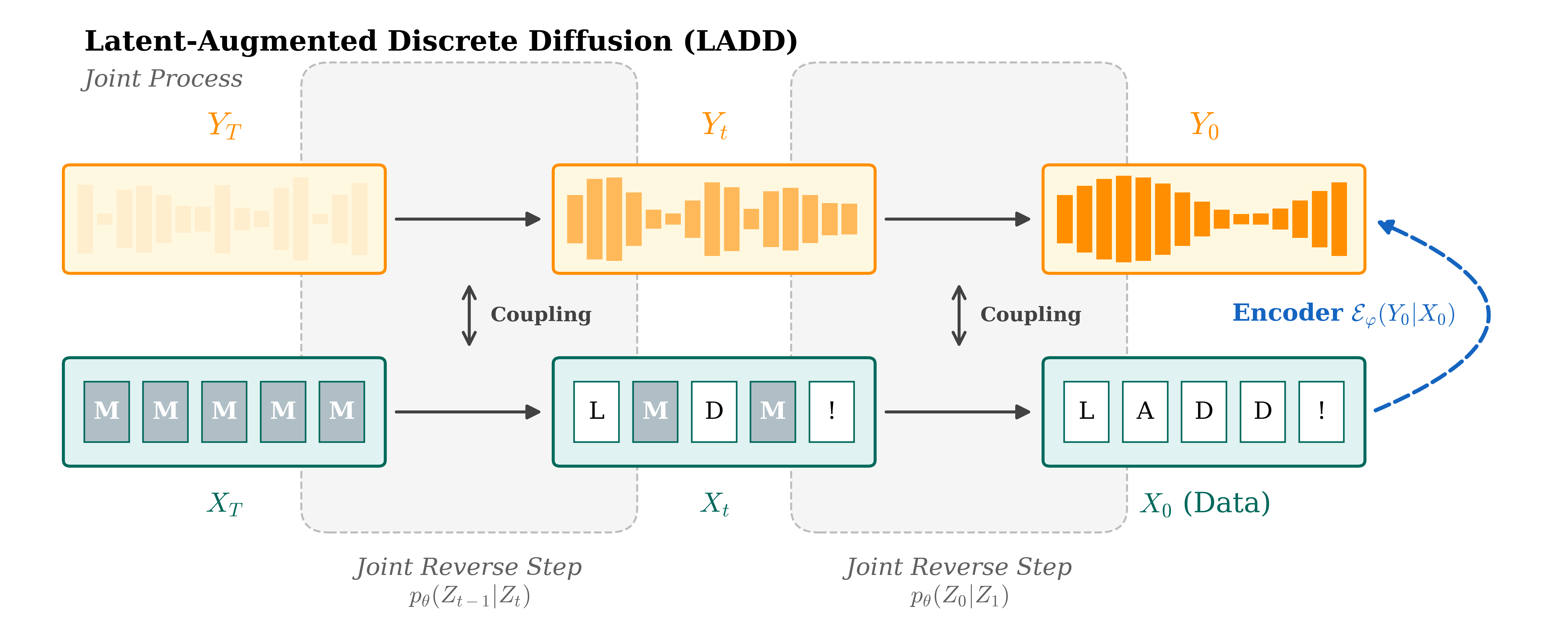}
    \caption{Overview of LADD (Joint Strategy). Tokens $X_t$ and auxiliary latents $Y_t$ are denoised simultaneously. The joint reverse process enables information flow between the latent channel and the data channel at every step, improving global coherence. An encoder $\mathcal{E}_\varphi$ provides the latents.}
    \label{fig:ladd_schematic}
\end{figure}

\section{Preliminaries on Diffusion Models}
\label{sec:ddms}
We first briefly present continuous and discrete diffusion models from a variational perspective.

\paragraph{Notation.}
$\mathcal{P}(\mse)$ denotes probability measures on $(\mse,\mathcal{E})$, 
$\KL(\mu\mid \nu)$ the Kullback--Leibler divergence, $\updelta_x$ the Dirac 
mass at $x$, and $\normal(\mu,\Sigma)$ the Gaussian with mean $\mu$ and 
covariance $\Sigma$. We write $\simplex_n$ for the probability simplex over 
$n$ outcomes and $\simplex_n^m$ for collections of $m$ such distributions.

\subsection{Continuous diffusion models}
\label{sec:ddpm}

We consider data $Y_0\sim \qy_0$ supported on $\latspace=\rset^{d}$. A 
standard discrete-time forward diffusion $\processd{Y_t}$ is defined by
\begin{equation}
\label{eq:forward_ddpm_transition}
Y_t = \alpha_t Y_{t-1} + \sigma_t \varepsilon_t\eqsp,
\end{equation}
with $\varepsilon_t\sim\normal(0,\Idd)$ i.i.d.\ and noise schedule 
$(\alpha_t,\sigma_t)$. The marginal admits the closed form 
$\q_{t\mid 0}(y_t \mid y_0) = \normal(y_t; \bar\alpha_t y_0, 
\bar\sigma_t^2 \Idd)$ with cumulative coefficients $\bar\alpha_t, \bar\sigma_t$ 
(\Cref{app:cont-diff}). We parameterize the reverse dynamics via a denoiser 
$\ypred(\cdot,t)$ with weights $\theta$ and train it under the $y_0$-prediction 
ELBO,
\begin{equation}
\label{eq:ddpm_loss}
\lossy(\theta) = \lossy_T + \lossy_0 + 
\mE\Big[ \sum_{t=2}^T \lambdayelbo_t \,\| Y_0 - \ypred(Y_t,t)\|^2 \Big]\eqsp,
\end{equation}
with $\lambdayelbo_t = \bar{\alpha}_{t-1}^2 \sigma_t^2 / 
(2\,\bar{\sigma}_{t-1}^2\,\bar{\sigma}_t^2)$. In practice, one uses a tunable 
$\lambday_t$ rather than the ELBO weight \citep{karras2022elucidating,
dieleman2024schedules}; we set $\lambday_t=1$ and use DDIM 
\citep{song2020denoising} for sampling.

\subsection{Masked Discrete Diffusion Models}
\label{sec:mdlm}

We consider data $\qx_0$ on sequences $\sspace^S$, where $\sspace=\{1,\ldots,K\}$ 
includes a dedicated mask state $\mask=K$. Let $(e_1,\ldots,e_K)$ be the 
canonical basis of $\rset^K$ and $\maskvec=e_K$. We identify each token with 
its one-hot vector and identify categorical distributions with points in the 
simplex $\simplex_K$. We focus on masked diffusion 
\citep{hoogeboom2021argmax,lou2024discrete,campbell2024generative,
sahoo2024simpleeffectivemaskeddiffusion}.

For $S=1$, the forward Markov chain starts from $X_0\sim\qx_0$ and evolves by
\begin{equation}
\label{eq:mdlm_markovian_forward_main}
\q_{t\mid t-1}(\cdot \mid x_{t-1}) =
\begin{cases}
\maskvec & \text{if } x_{t-1}=\maskvec\\
\gamma_t\,x_{t-1}+(1-\gamma_t)\,\maskvec & \text{otherwise}
\end{cases}\eqsp,
\end{equation}
with schedule $\gamma_t\in[0,1]$, $\bar\gamma_t=\prod_{s=1}^t\gamma_s$, and 
$\bar\gamma_T=0$ so that $X_T=\maskvec$ a.s. The marginal is 
$\q_{t\mid 0}(\cdot \mid x_0)=\bar\gamma_t\,x_0+(1-\bar\gamma_t)\,\maskvec$. 
For $S>1$ the process acts component-wise. The reverse model is initialized 
at the all-mask state and factorizes across positions:
\begin{align}
\label{eq:mdlm_generative_main}
\pthetax(x_{0:T})=\Big[\prod_{i=1}^S \updelta_{\maskvec}(x_T^i)\Big]\ \prod_{t=1}^T \pthetax_{t-1\mid t}(x_{t-1}\mid x_t)\eqsp, \quad \pthetax_{t-1\mid t}(x_{t-1}\mid x_t) =\prod_{i=1}^S \pthetaxi_{t-1\mid t}(x_{t-1}^i\mid x_t)\eqsp.
\end{align}
Let $\xpred(x_t,t)\in\simplex_K^S$ denote per-position categorical predictions 
with the mask coordinate set to zero. We parameterize $\pthetaxi_{t-1\mid t}$ 
by replacing the unknown $x_0^i$ in the exact bridge $\q_{t-1\mid t,0}$ with 
$\xpred^i(x_t,t)$ \citep{sahoo2024simpleeffectivemaskeddiffusion}; closed 
form in \Cref{app:discrete_diffusion}. The negative ELBO reduces to
\begin{align}
\label{eq:mdlm_discrete_nelbo_main}
\lossx(\theta) = \lossx_T + \lossx_0 + 
\mE\l[\sum_{t=2}^T \lambdaxelbo_t \sum_{i=1}^S 
-\mathbf{1}_{\{X_t^i=\maskvec\}}\, 
\log \big\langle \xpred^i(X_t,t),\,X_0^i \big\rangle \r],
\end{align}
with $\lambdaxelbo_t=(\bar\gamma_{t-1}-\bar\gamma_t)/(1-\bar\gamma_t)\geq 0$. 
The endpoint $\lossx_T$ vanishes when both forward and reverse are Dirac at 
the all-mask state; $\lossx_0$ is the one-step reconstruction loss. The 
factorization constraint in \eqref{eq:mdlm_generative_main} motivates the latent augmentations we introduce next.

\subsection{Limits and motivation}
\label{sec:limits-motivation}

Masked discrete diffusion models are typically trained with a reverse process 
that factorizes across positions, as in \eqref{eq:mdlm_generative_main}. This 
constrains the reverse dynamics to ignore cross-token couplings. The issue is 
most visible in the few-step regime: when many tokens are revealed at once, 
independent per-token decisions can combine into globally inconsistent 
sequences, since unmasking is a hard commitment, in contrast with the soft, 
reversible updates of continuous diffusion \citep{kang2025parallelbenchunderstandingtradeoffsparallel}.

Predictor-corrector samplers and 
tailored unmasking schedules help but do not fully remove the bias, and exhibit mild mode collapse and entropy–quality trade-offs 
\citep{wang2025remaskingdiscretediffusionmodels,gat2024discreteflowmatching,
liu2025thinkgeneratediscretediffusion,shi2024simplified,pham2025discrete}. The following proposition makes the limitation explicit by lower bounding the best achievable ELBO of any factorized reverse model.

\begin{mdframed}[style=propFrame]
\begin{proposition}{\citep[Proposition 1]{liu2025discretecopuladiffusion}}
\label{prop:factorization_limits}
Let $\qx$ be the path measure of the forward masked diffusion process on $\sspace^S$. Let $\pthetax$ be any factorized Markovian reverse model initialized at the all-mask state. Then
\begin{align}
\label{eq:factorization_lb}
\lossx(\theta)
\ \ge\ 
\entropy(\qx_0)
\  +\sum_{t=1}^T
\mE\Big[
\KL (
\qx_{t-1\mid t}(X_{t-1}\mid X_t)
 \Big|
\prod_{i=1}^S \qxi_{t-1\mid t}(X_{t-1}^i\mid X_t)
)
\Big]
\end{align}
where $\lossx$ denotes the negative ELBO (e.g., \eqref{eq:mdlm_discrete_nelbo_main}), $\entropy$ is the Shannon entropy, and $\qxi_{t-1\mid t}$ is the $i$-th marginal of the true reverse conditional $\qx_{t-1\mid t}$.
\end{proposition}
\end{mdframed}

Thus, any factorized reverse model is fundamentally limited by how far the true reverse conditionals deviate from a product form. The next proposition shows how conditioning on an auxiliary variable can tighten the lower bound, provided it reduces conditional dependence.

\begin{mdframed}[style=propFrame]
\begin{proposition}
\label{prop:latent_factorization_limits}
Let $(X_0,Y)\sim \qz_0$ with marginals $\qx_0$ and $\qy_0$.
Consider the masked forward diffusion on $X$ (independent of $Y$ given $X_0$) and denote by $\qx_{\mid y}$ the induced conditional path measure of $(X_{0:T}\mid Y=y)$.
Let $\qx_{t-1\mid t,y}(\cdot\mid x_t,y)$ be its (true) conditional reverse kernel and $\qxi_{t-1\mid t,y}$ its $i$-th marginal.
Suppose we model $X$ with a mixture of conditional reverse models,
\begin{equation}
\label{eq:latent_mixture_model}
\pthetax(x_{0:T})
=
\int \pthetax_{\mid y}(x_{0:T}\mid y)\ \qy_0(\dd y)\eqsp,
\end{equation}
where $\pthetax_{\mid y}(\cdot\mid y)$ is a factorized masked reverse process of the form \eqref{eq:mdlm_generative_main}, conditioned on $y$.
Then
\begin{equation}
\label{eq:latent_factorization_lb}
\lossx(\theta)
\ \ge\
\mE\big[\entropy(\qz_{0\mid y}(\cdot\mid Y))\big]
+\sum_{t=1}^T
\mE\Big[
\KL(
\qx_{t-1\mid t,y}(\cdot\mid X_t,Y)
\Big|
\prod_{i=1}^S \qxi_{t-1\mid t,y}(\cdot\mid X_t,Y)
)\Big]
\end{equation}
where the expectation is over $(X_0,Y)\sim \qz_0$ and $X_t\sim \qx_{t\mid 0}(\cdot\mid X_0)$.
\end{proposition}
\end{mdframed}

See \Cref{app:discrete_diffusion} for a proof. If $Y$ is such that $X$ becomes 
closer to conditionally factorized given $Y$ (and $\qz_{0\mid y}$ becomes lower entropy), the lower bound in \Cref{prop:latent_factorization_limits} tightens. 
In our approach, since $Y$ is not observed, we introduce an encoder $\enc_{\varphi}(\cdot\mid X)$ that produces the auxiliary latent, and we model its 
distribution with a separate latent diffusion process. This induces a trade-off between the expressivity of $Y$ and the difficulty of generating it, formalized in the next section.

\begin{remark}
This perspective also distinguishes latent augmentation from scaling the base 
masked diffusion model. Increasing denoiser capacity can sharpen per-token 
marginals, but does not change the product-form reverse kernel in 
\eqref{eq:mdlm_generative_main}: in the one-step regime, an optimal factorized 
denoiser still samples tokens independently from a fully masked state. With 
latent augmentation, an informative auxiliary variable can make the 
conditional posterior $\qxi_{t-1 |t, y}$ closer to factorized, so the one-step error can vanish.
\end{remark}

\section{Latent-Augmented Discrete Diffusion Models}
\label{sec:ladds}

We propose \emph{Latent-Augmented Discrete Diffusion} (LADD), which performs diffusion over an augmented state $Z=(X,Y)$, where $X\in\sspace^S$ is the token sequence and $Y$ is an auxiliary latent sequence intended to carry cross-token information. LADD retains parallel updates by factorizing transitions within each channel, while allowing cross-channel coupling through conditioning.
\subsection{Framework}
\label{sec:ladds_framework}

We consider two latent instantiations: \textbf{Co}ntinuous latents 
(\textbf{Co-LADD}) using a continuous diffusion process, and \textbf{Di}screte 
latents (\textbf{Di-LADD}) using a masked discrete diffusion process. 
Independently of the latent type, we use two \emph{process strategies}: a 
\emph{joint} strategy where tokens and latents are denoised together and 
interact at every reverse step (\Cref{fig:ladd_schematic}), and a 
\emph{sequential} strategy where the model first generates a clean latent and 
then denoises tokens conditionally on it. Both keep within-channel 
transitions factorized over positions.

The encoder $\enc_\varphi$ defines the joint distribution
\begin{equation}
\label{eq:qz0_def}
\qz_0(x_0,y_0)=\qx_0(x_0)\,\enc_\varphi(y_0\mid x_0)\eqsp.
\end{equation}
Let $M$ denote the latent length and $d_\ell$ the latent width. In Co-LADD, 
$\enc_\varphi=\normal(\mu_\varphi(x_0),\siglat^2 \mathrm{I}_{M d_\ell})$ with 
$\mu_\varphi(x_0)\in\rset^{M d_\ell}$; we fix $\siglat^2=10^{-4}$. 
Di-LADD applies a vector quantizer on top (\Cref{app:vq}). 

Sequences in $\sspace^S$ are said to lie in the data channel, sequences in $\latspace$ are said to lie in the latent channel.

\paragraph{Forward noising process.}
Starting from $Z_0\sim\qz_0$, we define a forward Markov chain $\processd{Z_t=(X_t,Y_t)}$ that factorizes across channels given $Z_0$:
\begin{equation}
\label{eq:ladds_forward_factorization}
\q(z_{0:T})
=
\qz_0(z_0)\,
\qx_{|0}(x_{1:T}\mid x_0)\,
\qy_{|0}(y_{1:T}\mid y_0)\eqsp,
\end{equation}
where $\qx_{|0}$ is the masked token forward process from \Cref{sec:mdlm}, and $\qy_{|0}$ is either Gaussian (Co-LADD; \Cref{sec:ddpm}) or masked discrete (Di-LADD), each with its own noise schedule.

\paragraph{Joint and sequential reverse models.}
The joint reverse model couples both channels at every step,
\begin{equation}
\tag{Joint}
\label{eq:ladds_joint_reverse}
\ptheta(z_{0:T}) = \ptheta_T(z_T)\,\prod_{t=1}^T \ptheta_{t-1\mid t}(z_{t-1}\mid z_t),
\end{equation}
with transitions
$\ptheta_{t-1\mid t}(x_{t-1},y_{t-1}\mid x_t,y_t) = 
\pthetax_{t-1\mid t}(x_{t-1}\mid x_t,y_t)\,
\pthetay_{t-1\mid t}(y_{t-1}\mid y_t,x_t)$ channel-factored and initialized at the all-mask token 
state. The sequential reverse model first samples a clean latent and then 
denoises tokens conditioned on it, allowing separate horizons $T^{\sspace}, 
T^{\latspace}$:
\begin{align}
\tag{Seq.}
\label{eq:ladds_seq_reverse}
\ptheta(x_{0:T^{\sspace}},y_{0:T^{\latspace}})
=
\pthetay_{T^{\latspace}}(y_{T^{\latspace}})
\prod_{t=1}^{T^{\latspace}}\pthetay_{t-1\mid t}(y_{t-1}\mid y_t)
\cdot
\pthetax_{T^{\sspace}}(x_{T^{\sspace}})
\prod_{t=1}^{T^{\sspace}}\pthetax_{t-1\mid t}(x_{t-1}\mid x_t,y_0)\eqsp,
\end{align}
with $y_0$ the final latent sample. Reverse kernels use the variational 
bridges of \Cref{sec:ddpm,sec:mdlm}; the token predictor outputs 
$\xpred(\cdot,t)\in\simplex_K^S$ and the latent predictor outputs 
$\ypred(\cdot,t)\in\simplex_C^M$ (Di-LADD) or $(\rset^{d_\ell})^M$ (Co-LADD). 
We use DDIM \citep{song2020denoising} for Co-LADD.

\subsection{Objective functions}
\label{sec:ladds_objectives}

We optimize the negative ELBO of the joint forward path measure $\q$ on 
$Z=(X,Y)$. Up to standard endpoint and encoder terms (derivations in 
\Cref{app:ladd_elbo}), it decomposes into a token loss $\lossx$ plus a latent 
loss $\lossy$. Define the conditioning variables 
$C_t^{\mathrm{joint}}=(X_t,Y_t)$ for joint inference, and 
$C_t^{\mathrm{seq},x}=(X_t,Y_0)$, $C_t^{\mathrm{seq},y}=Y_t$ for sequential 
inference.
The token reconstruction is the masked negative log-likelihood from 
\Cref{sec:mdlm}:
\begin{equation}
\label{eq:ladds_token_reconstruction}
\lossx = \mE\Bigg[
\sum_{t=2}^{T^{\sspace}} \lambda_t^x \sum_{i=1}^S 
-\mathbf{1}_{\{X_t^i=\maskvec\}}\,
\log \big\langle \xpred^i(C_t,t),\,X_0^i \big\rangle
\Bigg]\eqsp.
\end{equation}
For Co-LADD, the latent loss is the standard regression; for Di-LADD, the 
masked categorical:
\begin{equation}
\label{eq:ladds_latent_reconstruction}
\lossy_{\mathrm{Co}}
=
\mE\Bigg[
\sum_{t=2}^{T^{\latspace}} \lambda_t^y\,
\| Y_0 - \ypred(C_t,t)\|^2
\Bigg]\eqsp, \quad  \lossy_{\mathrm{Di}}
=
\mE\Bigg[
\sum_{t=2}^{T^{\latspace}} \lambda_t^y
\sum_{j=1}^M -\mathbf{1}_{\{Y_t^j=\mask\}}\,
\log \big\langle \ypred^j(C_t,t),\,Y_0^j \big\rangle
\Bigg]\eqsp.
\end{equation}
where $C_t$ uses the joint or sequential conditioning variables defined above.
We train the vector quantizer on top of the continuous encoder with a VQ commitment loss $\loss_{\mathrm{VQ}}(\varphi)$.
Optimization and codebook details are deferred to \Cref{app:vq}.
Putting the pieces together, we optimize
\begin{equation}
\label{eq:ladds_practical_objective}
\loss(\theta,\varphi)
=
\lossx
+
\lambdalatent\,\lossy
+
\mathbf{1}_{\{\mathrm{Di}\}}\ \betavq\,\loss_{\mathrm{VQ}}(\varphi)\eqsp,
\end{equation}
with hyperparameters $\lambdalatent, \betavq$.

\begin{remark} 
LADD is not specific to masked diffusion. It generalizes to other processes 
(e.g., uniform discrete diffusion) by changing the token forward kernel, the 
reverse bridge, and therefore the categorical reconstruction terms. The 
underlying issue of cross-token dependence under factorized updates remains, 
so we expect latent augmentation to help, although optimal schedules and 
parameterizations may differ.
\end{remark}

\subsection{Training procedure}
\label{sec:ladds_training}

Two design choices are intrinsic to LADD; we defer lower-level optimization details to \Cref{app:further_training_considerations}.

\paragraph{Two-stage training.} The LADD objective creates a tension: 
increasing the information content of $Y_0$ improves token denoising but 
makes the latent distribution harder to model. To decouple representation 
learning from latent generative modeling, we use a two-stage procedure: Stage~1 sets 
$\lambda_{\mathrm{latent}}=0$ and optimizes the token reconstruction and the vector quantizer loss, if used, then Stage~2 freezes the encoder and optimizes the full objective.

\paragraph{Joint and sequential inference from a single model.} 
We train the token denoiser to be robust to varied latent noise levels, so 
that the same model supports both inference schedules. For Di-LADD, we 
sample $t_x$ and $t_y$ independently. For Co-LADD, we use classifier-free 
latent dropout with probability $p_{\mathrm{cfg}}$, which we found more 
effective than independent timesteps.

\section{Related works}

We review concurrent advances in discrete diffusion that address the 
limitations of factorized denoisers. A more extensive review is in 
\Cref{app:related_works}, and we include a comparative table in \Cref{tab:related_work_comparison}. 

\paragraph{Non-factorized denoisers.}
One line of work augments masked diffusion with inference-time, non-factorized predictions. Discrete Copula Diffusion (DCD) \citep{liu2025discretecopuladiffusion} pairs per-token diffusion marginals with an auxiliary autoregressive copula to impose dependency structure, improving few-step quality at the cost of an AR loop. Self-speculative decoding \citep{campbell2025selfspeculativemaskeddiffusions} modifies attention to draft and verify a joint proposal in a single forward pass. These approaches operate purely within the discrete diffusion process and primarily target the sampler.

\paragraph{Continuous latents.}
The closest concurrent line of work introduces continuous variables into 
discrete diffusion. CCDD \citep{zhou2025coevolutionarycontinuousdiscretediffusion} 
co-evolves continuous states with discrete tokens; in our notation, it is a 
particular joint Co-LADD instantiation with token-aligned latents ($M=S$) and 
a frozen pretrained encoder. CADD \citep{zheng2025continuouslyaugmenteddiscretediffusion} 
couples encoder-derived continuous latents to token reconstruction without 
explicit latent modeling. VADD \citep{xie2025variationalautoencodingdiscretediffusion} 
introduces VAE-style continuous variables tied to each reverse step. 
Contrary to CCDD, CADD and VADD are not special cases of the LADD framework as we constructed it.

\paragraph{Latent diffusion for text.}
A related family performs diffusion in a learned latent space and decodes 
text from the final latent: TEncDM \citep{shabalin2025tencdmunderstandingpropertiesdiffusion}, 
COSMOS \citep{meshchaninov2026cosmoscompressedsmoothlatent}, 
LD4LG \citep{lovelace2023latentdiffusionlanguagegeneration}. Their pipeline is 
$X_0 \xrightarrow{\mathrm{enc}} Y_0 \xrightarrow{\mathrm{diff}} \hat Y_0 
\xrightarrow{\mathrm{dec}} \hat X_0$. LADD augments token diffusion rather 
than replacing it: $X_0 \xrightarrow{\mathrm{enc}} (X_0,Y_0) 
\xrightarrow{\mathrm{joint/seq.\ diff}} (\hat X_0,\hat Y_0)$, enabling 
multiple token-space refinement steps via the augmented $(X,Y)$ state space. 
The closest analog is sequential LADD, but the decoder differs: one-step 
non-autoregressive (COSMOS, TEncDM) or autoregressive (LD4LG).

\paragraph{Pretrained encoders.}
In the image domain, Representation Autoencoders \citep{zheng2025diffusiontransformersrepresentationautoencoders} 
advocate decoupling representation learning from generative modeling by 
pairing a frozen pretrained encoder with a learned latent diffusion model, 
a perspective we adopt in our text experiments via the Qwen3 encoder.

\section{Experiments}
\label{sec:exp}

We evaluate LADD in two settings. A controlled synthetic task 
(\Cref{sec:lowdim_exp}) isolates the conditionally factorized mechanism. Unconditional language 
modeling on LM1B (\Cref{sec:text_exp}) and OpenWebText 
(\Cref{sec:owt_exp}) then tests the design choices at scale: Co-LADD vs.\ Di-LADD, length-matched vs.\ compressed latents, and joint vs.\ sequential inference. 

Direct empirical comparison to concurrent 
latent-augmentation methods is nontrivial because reported results differ in 
tokenizer, training budget, code availability, and entropy calibration (Gen PPL can be artificially improved under diversity collapse \citep{pynadath2026generativefrontiersevaluationmatters}).
Thus, we focus our comparisons on a single strong masked discrete diffusion baseline, MDLM \citep{sahoo2024simpleeffectivemaskeddiffusion}, and include an AR reference on LM1B, trained 
under matched conditions (\Cref{tab:model_sizes_and_configs}). 
AR remains the strongest absolute-quality baseline, but LADD targets the parallel-generation frontier within the discrete-diffusion family.

LADD models use a multi-modal Diffusion Transformer (MM-DiT; \citep{esser2024scalingrectifiedflowtransformers}) 
; a plain shared-backbone DiT performed worse in preliminary experiments. MDLM uses a DiT. 

On text, we report data validation perplexity (data Val PPL), derived from the ELBO loss term $\lossx$, which is not the full ELBO for LADD variants; the goal is to confirm that learned latents are helpful for token reconstruction. Matched-compute comparisons report generative perplexity (Gen PPL) at fixed backbone-only GFLOP budgets; values are linearly interpolated between measured sampling budgets, and \texttt{n/a} is reported instead of extrapolation.

Further details and full
metric definitions are in 
\Cref{app:neural_network_architectures,app:metrics}.

\subsection{Controlled experiment: binary sawtooth}
\label{sec:lowdim_exp}

We use a controlled synthetic task to probe two questions: (i) can the auxiliary latent help in the few-step regime, and 
(ii) how is information carried as a function of latent type, discrete or continuous. We use the randomly shifted \emph{binary 
sawtooth} dataset \citep{pham2025discrete}, defined on $\{0,1\}^{S}$ by
\begin{equation}
\label{eq:binsaw_data_model_exp}
\qx_0=\qx_{0\mid y}\,\qy\eqsp,\quad 
\qx_{0\mid y}(x_0\mid y)=\prod_{i=1}^S \qxi_{0\mid y}(x_0^i\mid y)\eqsp,
\end{equation}
where $Y\sim \mathrm{U}[0,1]$ and $\qxi_{0\mid y}(\cdot\mid y)\sim 
\mathcal{B}(\omega^{\,i-y})$, with $\omega^{\,i-y}$ a $1$-periodic sawtooth 
ramp (explicit formula in \eqref{eq:omega_sawtooth} and illustration in
\Cref{fig:sawtooth_distribution}). By construction, $\qx_{0\mid y}$ factorizes while $\qx_0$ does not, and each marginal $\qxi_0$ is $\mathcal{B}(1/2)$.
We use $S=128$ and a single latent $M=1$; a continuous vector of width 
$d_\ell=32$ for Co-LADD, or a single code for Di-LADD. 
Further details in \Cref{app:exp_lowdim}.

\begin{figure}[t]
    \centering
    \begin{minipage}[t]{0.52\linewidth}
        \centering
        \caption{SWD$\downarrow$ versus sampling compute.}
        \includegraphics[width=0.85\linewidth]{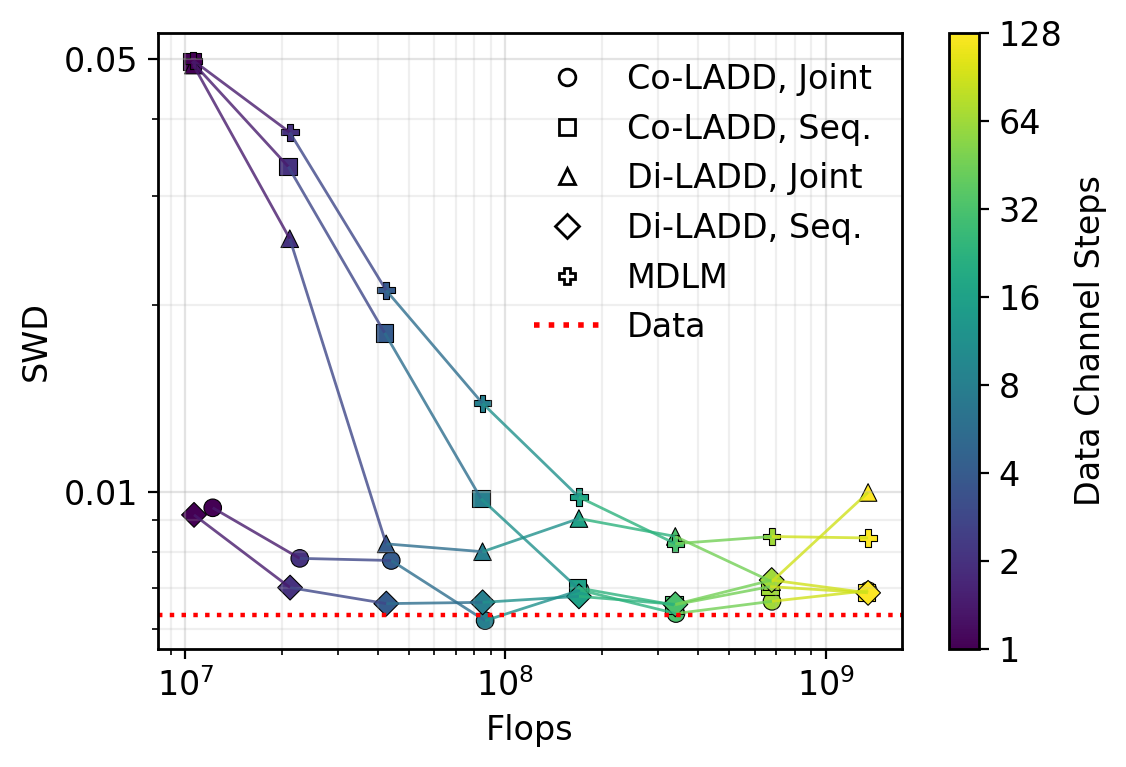}
        \label{fig:sawtooth}
    \end{minipage}
    \hfill
    \begin{minipage}[t]{0.45\linewidth}
        \centering
        \captionof{table}{
        Data reconstruction loss $\lossx$. \emph{Oracle} is the true conditional model in \eqref{eq:binsaw_data_model_exp}.
        }
        \label{tab:reconstruction_loss}
        \scalebox{0.8}{
        \begin{tabular}{lc}
            \toprule
            \textbf{Method} & \textbf{$\lossx$} \\
            \midrule
            \emph{Oracle} & 0.5075 \\
            MDLM & 0.5301 \\
            \midrule
            Di-LADD, Joint & 0.5055 \\
            Di-LADD, Seq. & 0.5067 \\
            Co-LADD, Joint & 0.4109 \\
            Co-LADD, Seq. & \textbf{0.2558} \\
            \bottomrule
        \end{tabular}
        }
    \end{minipage}

\end{figure}

\Cref{fig:sawtooth} shows that joint LADD improves over MDLM at small 
sampling budgets, and that sequential LADD reaches near-target Sliced Wasserstein Distance (SWD) with only 
one or two token denoising steps. This is the intended mechanism: once the 
latent resolves the global shift $y$, the conditional token distribution 
becomes approximately factorized, so many tokens can be unmasked coherently 
in parallel. MDLM, lacking this global signal, must resolve the same 
information over many steps.

\paragraph{Continuous latents carry more information, discrete codes extract sufficient statistic.}
\Cref{tab:reconstruction_loss} reports the token reconstruction loss 
$\lossx$. \emph{Oracle} conditions on the true shift $y$ and uses the known 
conditional marginals: it is therefore optimal among predictors that 
observe only $y$. Di-LADD matches this oracle, indicating that a single 
discrete code suffices to robustly capture the generative factor of variation. 
Co-LADD attains \emph{lower} reconstruction loss than the oracle. Since its encoder observes $X_0$, the 
continuous latent acts as a higher-bandwidth side channel, carrying 
sample-specific information beyond the latent shift. This improves token 
reconstruction but makes the latent distribution itself harder to model 
generatively. We will see this bandwidth contrast play out on text experiments.

\begin{remark}
\label{remark:no_seq_coladd}
In our text experiments, the bandwidth of the continuous latent makes its own diffusion model a bottleneck under sequential inference. We therefore omit Co-LADD (Seq.) in the following.
\end{remark}

\begin{figure}[t]
    \centering

    \begin{minipage}[t]{\textwidth}
        \centering
        \captionof{table}{
        LM1B Gen PPL$\downarrow$. Best is \textbf{bolded}, second best is \underline{underlined}, and third best is \emph{emphasized}.
        }
        \label{tab:lm1b-gen-ppl-matched-flops}
        \scalebox{0.8}{
        \begin{tabular}{lrrrrrrrr}
        \toprule
        GFLOPs & 50 & 100 & 200 & 400 & 800 & 1600 & 2500\\
        \midrule
        MDLM
        & 2724.60
        & 1032.90
        & 434.15
        & \emph{278.76}
        & \textbf{191.07}
        & \textbf{162.63}
        & \textbf{156.51}\\
        \midrule
        Co-LADD-32
        & \textbf{1180.20}
        & \textbf{524.22}
        & \underline{334.14}
        & 286.98
        & 224.64
        & {199.02}
        & \texttt{n/a} \\
        Di-LADD-32
        & \emph{1654.40}
        & \emph{656.97}
        & \emph{335.03}
        & \underline{256.67}
        & \underline{206.48}
        & \underline{187.32}
        & \texttt{n/a} \\
        Di-LADD-32 (SEQ)
        & \underline{1610.33}
        & \underline{558.00}
        & \textbf{299.67}
        & \textbf{236.13}
        & \emph{218.93}
        & 199.13
        & \texttt{n/a} \\
        \midrule 
        Co-LADD-128
        & 2977.99
        & 1145.40
        & 514.38
        & 324.44
        & 248.66
        & 197.70
        & \emph{198.43} \\
        Di-LADD-128
        & 2839.16
        & 1095.01
        & 468.19
        & 281.36
        & 220.81
        & \emph{191.91}
        & \underline{182.46} \\
        Di-LADD-128 (SEQ)
        & \texttt{n/a}
        & \texttt{n/a}
        & 421.48
        & 345.81
        & 317.62
        & 316.67
        & 302.33 \\
        \bottomrule
        \end{tabular}
        }
    \end{minipage}

\end{figure}

\subsection{Text modeling on LM1B}
\label{sec:text_exp}

We use LM1B to identify the design 
rules that we then validate at scale on OWT. We train all models for $300$k 
steps ($10$B tokens under classical accounting rules \citep{sahoo2024simpleeffectivemaskeddiffusion})
on packed sequences of length $S=128$ with the 
Qwen3 tokenizer, and match backbones at $184$M non-embedding parameters across 
MDLM, AR, and all LADD variants (\Cref{tab:model_sizes_and_configs}). 
The latent encoder is the frozen 
Qwen3-Embedding 0.6B model \citep{qwen3technicalreport}, so we only consider Stage 2 training. Further details are in \Cref{app:exp_lm1b}.

\begin{wraptable}[13]{r}{0.42\textwidth}
\centering
\vspace{-6pt}
\caption{
LM1B Data Val PPL$\downarrow$.
}
\label{tab:validation-128}
\scalebox{0.76}{
\begin{tabular}{lc}
\toprule
Method & Data Val PPL \\
\midrule
AR & 23.25 \\
\midrule
MDLM & $\leq 31.47$ \\
\midrule
Co-LADD-32 & $\leq 27.29$ \\
Di-LADD-32  & $\leq 25.29$ \\
Di-LADD-32, SEQ & $\leq 24.71$ \\
Co-LADD-128 & $\leq 23.66$ \\
Di-LADD-128 & $\leq 12.28$ \\
Di-LADD-128, SEQ & $\leq 12.08$ \\
\bottomrule
\end{tabular}
}
\vspace{-10pt}
\end{wraptable}

Using a frozen encoder is fully within the scope of LADD, whose purpose is exactly to exploit informative latent representations, pretrained or learned. 
We deliberately decouple representation learning from generative learning, as 
end-to-end encoder learning proved more delicate in preliminary experiments. 
As shown on the sawtooth task, a tailored encoder could improve factorizability further; we leave a systematic encoder study to future work.

All LADD variants improve over the baseline MDLM on data Val PPL, (\Cref{tab:validation-128}), showing that the latents are effectively informative. We mention again that Sequential Co-LADD is omitted (\Cref{remark:no_seq_coladd}), and we report some external references in \Cref{tab:lm1b_validation_all}.

\paragraph{LADD improves the compute-quality frontier}
\Cref{tab:lm1b-gen-ppl-matched-flops} reports Gen PPL at fixed compute budgets, and \Cref{tab:lm1b-entropy-matched-flops} shows entropies that are comparable. For reference, the AR baseline reaches a Gen PPL of 111.93\footnote{
Real data: Gen PPL 42.44 (entropy 4.40).
AR reference: Gen PPL 111.93 (entropy 4.41), temperature $\gamma=1.85$ to match data entropy.
}.
In the low-to-moderate compute regime, 
LADD dominates the frontier. These trends carry over to 
wall-clock time (\Cref{tab:lm1b_gen_ppl_entropy_steps_time_results_1}). At 
larger budgets ($\geq 1600$ GFLOPs), MDLM surpasses LADD, 
consistent with the asymptotic argument of \Cref{sec:limits-motivation}: as 
the number of steps grows, the factorization gap closes while LADD keeps paying the cost of modeling the latent channel.

\paragraph{Compressed latents win at matched compute.}
The $M=128$ variants improve data Val PPL clearly 
(\Cref{tab:validation-128}) but lose to MDLM at matched compute 
(\Cref{tab:lm1b-gen-ppl-matched-flops}): the latent channel adds 
per-step cost and is harder to model, so an over-long latent pays for itself on every sampling step, all while requiring more iterative refinements. 
Compressing to $M=32$ reduces per-step compute while preserving enough 
information to leverage, yielding the best matched-compute frontier.
Co-LADD-32 is best at the smallest budgets, achieving $522.22$ of Gen PPL where MDLM reaches $1032.90$, as its 
fine-grained continuous signal helps at low per-step cost. Di-LADD-32 takes over at moderate budgets, with its sequential version reaching a $30$\% improvement in Gen PPL over MDLM at $200$ GFlops and $16$\% at $400$ GFlops, with cleaner discrete latents aiding denoising.

\paragraph{Discrete latents yield an information tradeoff.}
Di-LADD's data Val PPL degrades when going from $M=128$ to $M=32$ (\Cref{tab:validation-128}), an early sign that the discretization becomes a bottleneck under compression, counterbalancing its role as a cleaner \emph{sufficient statistic} information channel. 
For the Di-LADD-32 sequential variant, we use $T^{\latspace}=8$, which yields improvements over its joint counterpart at small-to-moderate budgets, but, again, as the latent-modeling problem becomes harder, the right compute tradeoff is harder to find. These contrasts 
sharpen at OWT scale (\Cref{sec:owt_exp}).

\subsection{Language modeling on OpenWebText}
\label{sec:owt_exp}

We test whether the LM1B findings persist at larger scale. We train on 
OpenWebText (OWT) with sequence length $S=512$ for $500$k steps at batch 
size $512$ ($65$B tokens). Following the LM1B findings on compression, we focus on $M=16,64 < S$. Di-LADD-64 (SEQ) uses $8$ latent diffusion steps. Co-LADD emerges as the strongest configuration and serves as our representative LADD model.

\begin{table}[t]
\centering
\caption{
OWT zero-shot Data Val PPL$\downarrow$. Best \textbf{bolded}, second best \underline{underlined}, third best \emph{emphasized}.
}
\label{tab:owt_zeroshot}
\scalebox{0.8}{
    \begin{tabular}{lrrrrrrrr}
    \toprule
     & OWT & PTB & Wikitext & LM1B & Lambada & AG News & Pubmed & Arxiv \\
    \midrule
    MDLM
    & \emph{27.47}
    & \emph{88.60}
    & \emph{30.40}
    & 64.65
    & \emph{87.59}
    & \emph{56.67}
    & \emph{60.95}
    & \emph{61.60} \\
    \midrule 
    Co-LADD-16 & \underline{21.17} & \underline{78.11} & \underline{24.62} & \underline{50.30} & \underline{64.51} & \underline{47.24} & \underline{45.49} & \underline{48.41} \\
    Co-LADD-64
    & \textbf{20.74}
    & \textbf{59.11}
    & \textbf{22.61}
    & \textbf{48.66}
    & \textbf{59.98}
    & \textbf{44.95}
    & \textbf{43.30}
    & \textbf{44.82} \\
    Di-LADD-16 & 31.87 & 105.84 & 36.88 & 72.64 & 95.66 & 77.74 & 74.40 & 78.13 \\
    Di-LADD-64
    & 28.03
    & 99.13
    & 32.29
    & \emph{60.40}
    & 88.76
    & 63.35
    & 65.28
    & 68.46 \\
    \bottomrule
    \end{tabular}
}
\end{table}

\paragraph{LADD improves (zero-shot) data Val PPL and the compute--quality frontier.}
Compressed Co-LADD reduces OWT data Val PPL and transfers consistently better to seven held-out corpora (\Cref{tab:owt_zeroshot}), with $21$--$34\%$ relative reductions. This consistency across out-of-distribution corpora 
suggests the denoiser leverages latents beyond their OWT-specific features.
On the matched-compute frontier (\Cref{tab:owt-gen-ppl-matched-flops}), 
Co-LADD-16 and Co-LADD-64 reach Gen PPL $138.74$ and $185.16$ at $1600$ GFlops, versus MDLM's $298.40$, a resp. $53$\% and $38$\% reduction at iso-compute.
The advantage spans an $8\times$ compute window 
($\sim\!800$--$6400$ GFLOPs); entropy stays within $0.1$ nats of MDLM 
throughout, see \Cref{tab:owt-entropy-matched-flops}. The frontier gap is larger than on LM1B, consistent with longer sequences amplifying MDLM's factorization cost.
At sufficiently large budgets ($\geq 12800$ GFLOPs), MDLM eventually catches 
up, as the factorization gap closes while LADD keeps paying the latent 
modeling cost.

\begin{table}[ht!]
\centering
\caption{
OWT Gen PPL$\downarrow$. Best value \textbf{bolded}, second best \underline{underlined}, third best \emph{emphasized}.
}
\label{tab:owt-gen-ppl-matched-flops}
\scalebox{0.8}{

\begin{tabular}{lrrrrrr}
\toprule
GFlops & 800 & 1600 & 3200 & 6400 & 12800 & 25591 \\
\midrule
MDLM & 399.43 & 298.40 & \underline{132.03} & 113.17 & \textbf{82.38} & \textbf{81.73}\\
\midrule 
Co-LADD-16 & \underline{376.96} & \textbf{138.74} & \textbf{126.43} & \textbf{101.81} & \emph{97.88} & \underline{90.03} \\
Co-LADD-64 & \textbf{374.71} & \emph{185.16} & \emph{133.67} & \underline{104.10} & \underline{96.42} & \emph{92.60} \\
\midrule
Di-LADD-16 & \emph{377.42} & \underline{163.33} & 143.58 & \emph{104.11} & 111.17 & 125.29 \\
Di-LADD-64 & 467.08 & 212.93 & 148.17 & 118.05 & 117.08 & 128.41 \\
Di-LADD-16 (SEQ) & 415.57 & 196.26 & 166.82 & 113.30 & 106.93 & 94.46 \\
Di-LADD-64 (SEQ) & 404.17 & 242.43 & 190.13 & 155.67 & 144.91 & 136.91 \\
\bottomrule
\end{tabular}
}
\end{table}

\paragraph{Discrete latents struggle under their own bandwidth tradeoffs.}
Di-LADD-64 further illustrates the bandwidth tradeoff from the sawtooth experiment 
(\Cref{tab:reconstruction_loss}) and the LM1B results 
(\Cref{sec:text_exp}). Di-LADD is competitive with MDLM at best on zero-shot 
transfer and consistently behind Co-LADD on generative performance. Recovering competitiveness likely requires longer latents, larger codebooks, or hierarchical quantization. Sequential inference follows 
the same logic: Di-LADD (SEQ) is competitive only in narrow low-compute 
regimes, as the latent-generation problem becomes harder as the data distribution grows more complex.

Qualitative samples are given in \Cref{app:additional_results}.

\section{Conclusion}

We introduced \emph{Latent-Augmented Discrete Diffusion} (LADD), which augments masked token diffusion with an auxiliary latent channel and performs generation over the joint token--latent space.
The latent channel injects cross-token information while preserving parallel token updates, mitigating factored reverse transitions in few-step sampling.
Across controlled, LM1B, and OWT experiments, compressed LADD shifts the compute--quality frontier and positions latent augmentation as a practical route to faster non-autoregressive text generation.

\newpage

\begin{ack}
A.D. is funded by the European Union (ERC, Ocean, 101071601). D.S. and U.S. are partially funded by the European Union (ERC, Dynasty, 101039676). Views and opinions expressed are however those of the author(s) only and do not necessarily reflect those of the European Union or the European Research Council Executive Agency. Neither the European Union nor the granting authority can be held responsible for them.
U.S. is additionally funded by the French government under management of Agence Nationale de la Recherche as part of the ``France 2030'' program, reference ANR-23-IACL-0008 (PR[AI]RIE-PSAI).

The authors are grateful to the CLEPS infrastructure from the Inria of Paris for providing resources and support.
This work was granted access to the HPC resources of IDRIS under the allocation 2025-AD011015323R1 made by GENCI.
Part of this work was carried out while D.S. was an intern at Sakana AI.
\end{ack}

\bibliographystyle{plainnat}
\bibliography{reference}

\appendix

\newpage 
\section{Related works}
\label{app:related_works}

We review recent directions in discrete diffusion that reintroduce cross-token dependencies and/or couple the diffusion path with a continuous representation to improve performance. Classical ways to reintroduce dependencies include (i) autoregressive/Gibbs updates, (ii) explicit couplings (e.g., copulas or energy-based corrections), or (iii) latent variables that mediate global structure while retaining parallel token updates. We focus on the third option.

\paragraph{Non-factorized denoisers via copulas and speculative decoding.}
Discrete Copula Diffusion (DCD) \citep{liu2025discretecopuladiffusion} augments a discrete diffusion model at inference time with a separate copula model instantiated as a small autoregressive (AR) language model. At each step $t$, DCD combines (a) univariate marginals from causal and non-causal denoisers with (b) a dependency structure from the AR copula via a projection step. Sampling performs an AR pass over the currently masked subset followed by the diffusion update. Reported results show improved few-step quality with fewer calls to the diffusion model, at the cost of an auxiliary AR model and an additional AR pass whose compute scales with the masked span.

Self-speculative decoding \citep{campbell2025selfspeculativemaskeddiffusions} modifies the attention mask to include a causal drafting block and validates the draft against the full (non-causal) network via speculative acceptance. This yields a non-factorized proposal over the masked subset in a single forward of the full model and can reduce the number of evaluations in their setup.

\paragraph{Latent variables for dependency modeling.} 
\citet{xie2025variationalautoencodingdiscretediffusion} introduce Variational Autoencoding Discrete Diffusion (VADD). They associate one continuous latent $y_t$ per transition, obtained with an encoder $\enc_t^{\varphi}$ trained as part of a classical VAE, and sample a Gaussian latent at each transition during generation. This yields non-factorized posteriors and improves few-step quality without external AR models.

In contrast, our LADD formulation introduces a persistent latent channel that is coupled to the token diffusion through training objectives and inference-time conditioning. In particular, we explicitly model the latent channel (via its own diffusion/denoising mechanism), which supports stronger and more controllable conditioning than step-local Gaussian latents while keeping the token denoiser parallel.

\paragraph{Continuous augmentations.}
Contemporaneous work \citep{zhou2025coevolutionarycontinuousdiscretediffusion} proposes CCDD, which co-evolves continuous states attached to each discrete token. They do not explore variable latent sequence length $M$, end-to-end encoder learning regimes, or discrete latent tokens.

Contemporaneously, Continuously Augmented Discrete Diffusion (CADD) \citep{zheng2025continuouslyaugmenteddiscretediffusion}, attaches a continuous latent to each token position so that masked tokens carry soft hints instead of a hard $\mask$ token. Their design ties the continuous pathway closely to token reconstruction: they do not use a separate network for the latent denoiser, which they approximate coarsely from the data denoiser $\xpred$ and the encoder $\enc_{\varphi}$, simplifying training (data reconstruction loss only), but inducing important quality/entropy tradeoffs at inference. This also constrains how independently one can tune the latent channel versus the discrete channel at inference. Our focus is complementary: using a compact latent channel to encode global structure can be advantageous when many tokens are unveiled simultaneously or when long-range constraints dominate.

\paragraph{Using pre-trained encoders and joint feature modeling.}
In the image domain, Representation Autoencoders (RAE) \citep{zheng2025diffusiontransformersrepresentationautoencoders} advocate separating representation learning from generative modeling by using a fixed pretrained representation encoder, training a diffusion model on the resulting latent distribution, and training a decoder to perform inversion to the data domain. This perspective motivates considering powerful pretrained encoders for complex data distribution rather than learning one from scratch and specifically targeted at generative modeling.

\citet{kouzelis2026boostinggenerativeimagemodeling} boost generation by jointly synthesizing images and continuous features under coupled diffusion dynamics. While the modality and architectural choices differ, this line of work reinforces the broader theme that explicitly modeling a continuous side channel jointly with the discrete (or pixel-space) generation process can improve sample quality.

\paragraph{Latent diffusion for language generation.}
LD4LG \citep{lovelace2023latentdiffusionlanguagegeneration}, COSMOS \citep{meshchaninov2026cosmoscompressedsmoothlatent}, and TEncDM \citep{shabalin2025tencdmunderstandingpropertiesdiffusion} are representative latent-diffusion approaches for text generation. They share the general pipeline
\begin{equation}
    X_0
    \xrightarrow{\mathrm{enc}}
    Y_0
    \xrightarrow{\mathrm{diffusion}}
    \hat Y_0
    \xrightarrow{\mathrm{decoder}}
    \hat X_0 ,
\end{equation}
where diffusion is performed in a continuous representation space rather than directly on the discrete token sequence. LD4LG performs language generation by first mapping text to a continuous latent representation, learning a diffusion model over this latent space, and decoding generated latents back into text. COSMOS follows the same latent-diffusion principle, but emphasizes compressed sequence-level representations, so that generation proceeds through a lower-dimensional latent sequence before being decoded into tokens. TEncDM instead uses contextual token encodings as the diffusion state, applying diffusion in the space of encoder features and recovering text from the final denoised representation.

These methods are closely related in spirit to LADD, since they exploit continuous representations to make text generation more compatible with diffusion. The key difference is that their latent representation replaces token diffusion, whereas LADD augments token diffusion. In LADD, the generative process is defined on the joint state $(X,Y)$, or sequentially on $Y$ followed by $X\mid Y$, so token-space denoising remains an explicit part of generation:
\begin{equation}
    X_0
    \xrightarrow{\mathrm{enc}}
    (X_0,Y_0)
    \xrightarrow{\mathrm{joint/sequential\ diffusion}}
    (\hat X_0,\hat Y_0).
\end{equation}
Consequently, LD4LG, COSMOS, and TEncDM rely on a terminal decoder to map the final latent sample back to text, while LADD can use the generated latent to guide multiple subsequent token-refinement steps. This distinction is most visible in the sequential variant: although LADD-SEQ first resolves the latent channel, it still generates text through a conditional masked diffusion process rather than through a one-shot latent-to-text decoder.

We provide a more explicit comparison between LADD and closely related latent-augmentation or latent-diffusion methods in \Cref{tab:related_work_comparison}.

\begin{remark}
Direct empirical comparison across these methods is delicate. Reported results often differ in tokenizer, dataset, model size, training budget, sampling budget, entropy calibration, and code availability. Moreover, generative perplexity can be improved by reducing sample entropy, so comparisons that do not report or control entropy can be misleading. For this reason, our main experiments emphasize controlled comparisons against MDLM under matched training and evaluation conditions, and report sample entropy alongside Gen PPL.
\end{remark}

\begin{table}[t]
\centering
\caption{
Comparison of LADD with related methods. ``Token diffusion'' indicates whether generation includes an explicit diffusion process over tokens. ``Latent modeling'' indicates whether the latent is separately modeled. ``Token refinements'' indicates whether token-space denoising is available while or after latent information is produced.
}
\label{tab:related_work_comparison}
\resizebox{\textwidth}{!}{
\begin{tabular}{lcccc}
\toprule
Method
& Token diffusion
& Latent modeling
& Token refinements
& Relation to LADD \\
\midrule
DCD \citep{liu2025discretecopuladiffusion}
& Yes
& No
& Yes
& Non-factorized token sampler \\
Self-speculative MDLM \citep{campbell2025selfspeculativemaskeddiffusions}
& Yes
& No
& Yes
& Non-factorized denoising through proposal mechanism \\
VADD \citep{xie2025variationalautoencodingdiscretediffusion}
& Yes
& Step-local
& Yes
& VAE-style auxiliary variables for reverse transitions \\
CCDD \citep{zhou2025coevolutionarycontinuousdiscretediffusion}
& Yes
& Yes
& Yes
& Special case of joint Co-LADD ($M=S$, pretrained encoder) \\
CADD \citep{zheng2025continuouslyaugmenteddiscretediffusion}
& Yes
& No/Implied
& Yes
& Continuous augmentation, latents not modeled explicitly (Monte-Carlo) \\
TEncDM \citep{shabalin2025tencdmunderstandingpropertiesdiffusion}
& No
& Yes
& No
& Latent diffusion followed by terminal text decoder \\
COSMOS \citep{meshchaninov2026cosmoscompressedsmoothlatent}
& No
& Yes
& No
& Compressed latent diffusion followed by terminal text decoder \\
LD4LG \citep{lovelace2023latentdiffusionlanguagegeneration}
& No
& Yes
& No
& Latent diffusion for language generation \\
\midrule
LADD-SEQ
& Yes
& Yes
& Yes
& Latent diffusion followed by conditional token diffusion \\
LADD-JOINT
& Yes
& Yes
& Yes
& Joint diffusion on the augmented state $(X,Y)$ \\
\bottomrule
\end{tabular}
}
\end{table}

\section{Additional remarks on continuous and discrete diffusion}

\subsection{Derivations for continuous diffusion}
\label{app:cont-diff}

We collect standard identities for discrete-time Gaussian diffusion models (DDPM/DDIM style), following \citet{ho2020denoisingdiffusionprobabilisticmodels,song2020denoising}.

\paragraph{Forward process and marginals.}
Let $(\alpha_t,\sigma_t)_{t=1}^T$ be a noise schedule with $\alpha_t>0$ and $\sigma_t>0$, and consider the forward Markov chain
\begin{equation}
\label{eq:app_forward_step}
Y_t=\alpha_t\,Y_{t-1}+\sigma_t\,\varepsilon_t\eqsp,\qquad \varepsilon_t\sim\normal(0,\Idd)\ \text{i.i.d.}\eqsp,
\end{equation}
initialized at $Y_0\sim \qy_0$. Define cumulative coefficients by $\bar\alpha_0=1$, $\bar\sigma_0=0$, and for $t\ge 1$,
\begin{equation}
\label{eq:app_cumulatives_rec}
\bar\alpha_t=\prod_{s=1}^t \alpha_s\eqsp,
\qquad
\bar\sigma_t^2=\alpha_t^2\,\bar\sigma_{t-1}^2+\sigma_t^2\eqsp.
\end{equation}
Then for each $t\in\{0,\ldots,T\}$, the marginal conditional is
\begin{align}
\label{eq:app_forward_marginal}
\qy_{t\mid 0}(y_t\mid y_0)&=\normal\!\big(y_t;\ \bar\alpha_t\,y_0,\ \bar\sigma_t^2\,\Idd\big)\eqsp.
\end{align}
More generally, for any $0\le s<t\le T$, define
\begin{equation}
\label{eq:app_skip_defs}
\alpha_{t\mid s}=\frac{\bar\alpha_t}{\bar\alpha_s}\eqsp,
\qquad
\sigma_{t\mid s}^2=\bar\sigma_t^2-\alpha_{t\mid s}^2\,\bar\sigma_s^2\eqsp,
\end{equation}
so that the forward kernel is Gaussian:
\begin{equation}
\label{eq:app_forward_kernel}
\qy_{t\mid s}(y_t\mid y_s)=\normal\!\big(y_t;\ \alpha_{t\mid s}\,y_s,\ \sigma_{t\mid s}^2\,\Idd\big)\eqsp.
\end{equation}
The definitions in \eqref{eq:app_skip_defs} reduce to $\alpha_{t\mid t-1}=\alpha_t$ and $\sigma_{t\mid t-1}=\sigma_t$.

\paragraph{Bridge posteriors.}
For $0\le s<t\le T$, the bridge posterior $\qy_{s\mid t,0}(y_s\mid y_t,y_0)$ is Gaussian:
\begin{align}
\label{eq:app_bridge}
\qy_{s\mid t,0}(y_s\mid y_t,y_0)
&=\normal\!\big(y_s;\ \tilde\mu_{s\mid t}(y_t,y_0),\ \tilde\sigma_{s\mid t}^2\,\Idd\big)\eqsp,\\
\label{eq:app_bridge_params}
\tilde\mu_{s\mid t}(y_t,y_0)
&=\frac{\alpha_{t\mid s}\,\bar\sigma_s^2}{\bar\sigma_t^2}\,y_t+\frac{\bar\alpha_s\,\sigma_{t\mid s}^2}{\bar\sigma_t^2}\,y_0\eqsp,
\\ \tilde\sigma_{s\mid t}^2 &=\frac{\sigma_{t\mid s}^2\,\bar\sigma_s^2}{\bar\sigma_t^2}\eqsp.
\end{align}
Writing the (true) noise at time $t$ as
\begin{equation}
\label{eq:app_true_noise}
\varepsilon(y_t,y_0)=\frac{y_t-\bar\alpha_t\,y_0}{\bar\sigma_t}\eqsp,
\end{equation}
the posterior mean admits the equivalent form
\begin{equation}
\label{eq:app_bridge_eps_form}
\tilde\mu_{s\mid t}(y_t,y_0)=\frac{1}{\alpha_{t\mid s}}
\Big(y_t-\frac{\sigma_{t\mid s}^2}{\bar\sigma_t}\,\varepsilon(y_t,y_0)\Big)\eqsp.
\end{equation}

\paragraph{Path decomposition.}
The joint density of the forward trajectory factorizes as
\begin{equation}
\label{eq:app_path_decomp}
\qy_{0:T}(y_{0:T})
=\qy_0(y_0)\ \qy_{T\mid 0}(y_T\mid y_0)\ \prod_{t=2}^T \qy_{t-1\mid t,0}(y_{t-1}\mid y_t,y_0)\eqsp,
\end{equation}
where $\qy_{t-1\mid t,0}$ is given by \eqref{eq:app_bridge}--\eqref{eq:app_bridge_params} with $(s,t)\leftarrow(t-1,t)$.

\paragraph{ELBO decomposition.}
Let $\pthetay_T(y_T)=\normal(0,\bsigma_T^2\,\Idd)$ and define a Markov reverse model
\begin{equation}
\label{eq:app_reverse_family}
\pthetay(y_{0:T})=\pthetay_T(y_T)\prod_{t=1}^T \pthetay_{t-1\mid t}(y_{t-1}\mid y_t)\eqsp.
\end{equation}
The standard variational identity yields the negative ELBO
\begin{align}
\label{eq:ddpm_elbo_appendix}
\lossy(\theta)&=\mE\Big[
\underbrace{\KL\big(\qy_{T\mid 0}(\cdot\mid Y_0)\,\|\,\pthetay_T\big)}_{L_T}
\\ &+\sum_{t=2}^T \underbrace{\KL\big(\qy_{t-1\mid t,0}(\cdot\mid Y_t,Y_0)\,\|\,\pthetay_{t-1\mid t}(\cdot\mid Y_t)\big)}_{L_{t-1}}
\\ &\;-\;\underbrace{\log \pthetay_{0\mid 1}(Y_0\mid Y_1)}_{L_0}
\Big]\eqsp,
\end{align}
where the expectation is over $Y_0\sim\qy_0$ and $Y_t\sim\qy_{t\mid 0}(\cdot\mid Y_0)$.

\paragraph{From KL terms to weighted regression.}
A common parameterization matches the bridge covariance and replaces $y_0$ by a predictor:
\begin{align}
\label{eq:app_reverse_param}
\pthetay_{t-1\mid t}(\cdot\mid y_t)
&=\normal\!\big(\cdot;\ \tilde\mu_{t-1\mid t}(y_t,\ \hat y_0(y_t,t)),\ \tilde\sigma_{t-1\mid t}^2\,\Idd\big)\eqsp,
\\ \hat y_0(y_t,t)&=\ypred(y_t,t)\eqsp,
\end{align}
with $(s,t)\leftarrow(t-1,t)$ in \eqref{eq:app_bridge_params}. Since the two Gaussians in $L_{t-1}$ share the same covariance, $L_{t-1}$ reduces (up to an additive constant) to a weighted squared error in the prediction target. For $y_0$-prediction, one obtains
\begin{align}
\label{eq:app_y0_pred_weight}
L_{t-1}=\mE\Big[&\lambdayelbo_t\ \|Y_0-\ypred(Y_t,t)\|^2\Big]\ +\ \text{const}\eqsp,
\\ &\lambdayelbo_t=\frac{\bar\alpha_{t-1}^2\,\sigma_t^2}{2\,\bar\sigma_{t-1}^2\,\bar\sigma_t^2}\eqsp,
\end{align}
which matches the weight used in the main text. For $\varepsilon$-prediction, using \eqref{eq:app_bridge_eps_form} and a network $\varepsilon_\theta(y_t,t)$, the corresponding ELBO weight is
\begin{equation}
\label{eq:app_eps_pred_weight}
L_{t-1}=\mE\Big[\frac{\sigma_t^2}{2\,\bar\sigma_{t-1}^2\,\alpha_t^2}\ \|\varepsilon_\theta(Y_t,t)-\varepsilon(Y_t,Y_0)\|^2\Big]\ +\ \text{const}\eqsp,
\end{equation}
where $\varepsilon(Y_t,Y_0)$ is given by \eqref{eq:app_true_noise}. The two parameterizations are linked by
\begin{equation}
\label{eq:app_eps_y0_link}
Y_0=\frac{Y_t-\bar\sigma_t\,\varepsilon(Y_t,Y_0)}{\bar\alpha_t}\eqsp,
\qquad
\hat y_0(y_t,t)=\frac{y_t-\bar\sigma_t\,\varepsilon_\theta(y_t,t)}{\bar\alpha_t}\eqsp.
\end{equation}
In practice, the exact ELBO weights (e.g., $\lambdayelbo_t$) are often modified (e.g., constant, SNR-based, or schedule-based reweightings) without changing the model class \citep{karras2022elucidating,dieleman2024schedules}. Endpoint terms $L_T$ and $L_0$ are also typically small and often omitted.

\paragraph{Noise schedules.}
The relation between $(\bar\alpha_t,\bar\sigma_t)$ depends on the diffusion type. In the variance-preserving (VP) case, $\bar\sigma_t^2=1-\bar\alpha_t^2$. In the variance-exploding (VE) case, $\bar\alpha_t=1$ and $\bar\sigma_t^2=\sigma^2(t)$ with $\sigma^2(T)\gg 1$.
We focus on VP and consider $\tau=t/T\in[0,1]$. We list noise schedules in \Cref{tab:noise_schedules}. 

\paragraph{Deterministic DDIM sampling.}
DDIM \citep{song2020denoising} defines an alternative (possibly deterministic) sampler that preserves the same marginals while altering the reverse-time conditional.
For any $0\le s<t\le T$, define $\tilde\sigma_{s\mid t}^2$ as in \eqref{eq:app_bridge_params} and let $\eta\in[0,1]$ control stochasticity.
Given $y_t$, form $\hat y_0=\ypred(y_t,t)$ (or equivalently via $\varepsilon_\theta$ using \eqref{eq:app_eps_y0_link}) and set
\begin{equation}
\label{eq:app_ddim_kernel}
\ptheta_\eta(y_s\mid y_t)
=\normal\!\big(y_s;\ \tilde\mu^\eta_{s\mid t}(y_t,\hat y_0),\ \eta^2\,\tilde\sigma_{s\mid t}^2\,\Idd\big)\eqsp,
\end{equation}
with mean
\begin{align}
\label{eq:app_ddim_mean}
\tilde\mu^\eta_{s\mid t}(y_t,\hat y_0)
&=\bar\alpha_s\,\hat y_0
+\sqrt{\bar\sigma_s^2-\eta^2\,\tilde\sigma_{s\mid t}^2}\ \hat\varepsilon\eqsp,
\\ \hat\varepsilon &=\frac{y_t-\bar\alpha_t\,\hat y_0}{\bar\sigma_t}\eqsp.
\end{align}
Setting $\eta=1$ recovers the usual stochastic reverse consistent with the Markovian diffusion, while $\eta=0$ yields a deterministic trajectory (DDIM) that preserves the same one-time marginals. Training is unchanged; DDIM only modifies the sampling procedure.

\subsection{Discrete diffusion}
\label{app:discrete_diffusion}

We collect standard identities for discrete-time masked diffusion \citep{sahoo2024simpleeffectivemaskeddiffusion} and prove \Cref{prop:latent_factorization_limits}.

\paragraph{Forward process and marginals.}
For a single position, the forward kernel is
\begin{align}
\label{eq:app_mdlm_forward}
\q_{t\mid t-1}(\cdot\mid x_{t-1})
=
\begin{cases}
\maskvec & \text{if } x_{t-1}=\maskvec\\
\gamma_t\,x_{t-1}+(1-\gamma_t)\,\maskvec & \text{otherwise}
\end{cases}\eqsp,
\end{align}
with $\bar\gamma_t=\prod_{s=1}^t\gamma_s$. It follows that
\begin{equation}
\label{eq:app_mdlm_marginal}
\q_{t\mid 0}(\cdot \mid x_0)=\bar\gamma_t\,x_0+(1-\bar\gamma_t)\,\maskvec\eqsp.
\end{equation}
We choose $\bar\gamma_T=0$, hence $X_T=\maskvec$ almost surely. For sequences, the chain evolves independently across positions under the forward process.

\paragraph{Reverse bridge (closed form).}
For one position, the exact reverse bridge conditioned on $x_0$ satisfies
\begin{align}
\label{eq:app_mdlm_bridge_single}
\q_{t-1\mid t,0}(\cdot\mid x_t,x_0)
=
\begin{cases}
x_t & \text{if } x_t\neq \maskvec\\[2mm]
\displaystyle
\frac{1-\bar\gamma_{t-1}}{1-\bar\gamma_t}\,\maskvec
+\frac{\bar\gamma_{t-1}-\bar\gamma_t}{1-\bar\gamma_t}\,x_0
& \text{if } x_t=\maskvec
\end{cases}\eqsp.
\end{align}
For sequences, the bridge factorizes across positions:
\begin{equation}
\label{eq:app_mdlm_bridge_seq}
\q_{t-1\mid t,0}(x_{t-1}\mid x_t,x_0)=\prod_{i=1}^S \q_{t-1\mid t,0}(x_{t-1}^i\mid x_t^i,x_0^i)\eqsp.
\end{equation}

\paragraph{Generative parameterization.}
Let $\xpred(x_t,t)\in\simplex_K^S$ with $(\xpred^i(x_t,t))_K=0$.
Define the reverse transition by replacing $x_0^i$ with $\xpred^i(x_t,t)$ in the bridge:
\begin{align}
\label{eq:app_mdlm_param}
\pthetaxi_{t-1\mid t}(\cdot\mid x_t)
&=\q_{t-1\mid t,0}\big(\cdot\mid x_t^i,\ \xpred^i(x_t,t)\big)\eqsp, \\ 
\pthetax_{t-1\mid t}(x_{t-1}\mid x_t)&=\prod_{i=1}^S \pthetaxi_{t-1\mid t}(x_{t-1}^i\mid x_t)\eqsp.
\end{align}

\paragraph{ELBO and weighted cross-entropy.}
For any reverse model $\pthetax$ initialized at the all-mask state,
the negative ELBO admits the standard decomposition
\begin{align}
\label{eq:app_mdlm_elbo_decomp}
-\mE[\log \ptheta_0(X_0)] \leq  \lossx(\theta)
&=\mE\Bigg[
\underbrace{\KL\big(\q_{T\mid 0}(\cdot\mid X_0)\,\|\,\pthetax_T\big)}_{\lossx_T}
\\ &+\sum_{t=2}^T \underbrace{\KL\big(\q_{t-1\mid t,0}(\cdot\mid X_t,X_0)\,\|\,\pthetax_{t-1\mid t}(\cdot\mid X_t)\big)}_{\lossx_{t-1}}
\\ &-\underbrace{\log \pthetax_{0\mid 1}(X_0\mid X_1)}_{\lossx_0}
\Bigg]\eqsp,
\end{align}
where the expectation is over $X_0\sim\qx_0$ and $X_t\sim\q_{t\mid 0}(\cdot\mid X_0)$.
Under the parameterization \eqref{eq:app_mdlm_param}, only masked positions contribute to $\lossx_{t-1}$, and one obtains
\begin{align}
\label{eq:app_mdlm_elbo_ce}
&\lossx_{t-1}
=\mE\Bigg[
\lambdaxelbo_t
\sum_{i=1}^S -\mathbf{1}\{X_t^i=\maskvec\}\,
\log \big\langle \xpred^i(X_t,t),\,X_0^i\big\rangle
\Bigg]\\
\label{eq:app_mdlm_lambda}
& \qquad \qquad \qquad \quad \lambdaxelbo_t
=\frac{\bar\gamma_{t-1}-\bar\gamma_t}{1-\bar\gamma_t}\eqsp.
\end{align}
The endpoint term $\lossx_T$ is $0$ when both $\q_{T\mid 0}$ and $\pthetax_T$ are Dirac at the all-mask state.
The reconstruction term $\lossx_0$ reduces to
\begin{equation}
\label{eq:app_mdlm_L0}
\lossx_0
=
\mE\Bigg[
\sum_{i=1}^S -\mathbf{1}\{X_1^i=\maskvec\}\,
\log \big\langle \xpred^i(X_1,1),\,X_0^i\big\rangle
\Bigg]\eqsp,
\end{equation}
up to constants from positions that are already revealed at $t=1$.
As in continuous diffusion, the ELBO weights can be replaced by alternative reweightings without changing the model class.

\begin{proof}[Proof of \Cref{prop:latent_factorization_limits}]
Consider the augmented process $\Tilde{Z}_t=(X_t,Y_0)$ where $(X_0,Y_0)\sim \qz_0$, with $\qz_0(x, y)=\qz_{0 | y}(x|y)\qy_0(y)$, and $(X_t)_{t\ge 1}$ is driven by the forward masked process
\eqref{eq:mdlm_markovian_forward_main} conditionally on $X_0$.
Denote by $\qtildez$ the resulting path measure of $\processd{\Tilde{Z}_t}$.

\paragraph{Step 1: reduction to conditional likelihoods.}
Define the joint model at time $0$ induced by the mixture construction:
\begin{equation}
\label{eq:app_ptilde0}
\ptilde_{\theta,0}(x, y)=\qy_0( y)\,\pthetax_{0\mid y}( x\mid y)\eqsp.
\end{equation}
Since $\qz_0$ and $\ptilde_{\theta,0}$ share the same marginal $\qy_0$ on $Y$, the KL chain rule gives
\begin{equation}
\label{eq:app_kl_chain_rule_userstyle}
\begin{aligned}
\KL\big(\qz_0\ \big|\ \ptilde_{\theta,0}\big)
&= \mE\Big[\KL\big(\qz_{0\mid y}(\cdot\mid Y)\ \big|\ \pthetax_{0\mid y}(\cdot\mid Y)\big)\Big]
\\
&= \mE\Big[\log \qz_{0\mid y}(X_0\mid Y)-\log \pthetax_{0\mid y}(X_0\mid Y)\Big]\eqsp.
\end{aligned}
\end{equation}
In particular,
\begin{align}
\label{eq:app_neglog_decomp}
\mE\big[-\log \pthetax_{0\mid y}(X_0\mid Y)\big]
&=
\mE\big[\entropy(\qz_{0\mid y}(\cdot\mid Y))\big]
+
\KL\big(\qz_0\mid \ptilde_{\theta,0}\big)
\end{align}

\paragraph{Step 2: ELBO on the conditional diffusion path.}
Consider the conditional reverse model $\pthetax_{\mid y}(x_{0:T}\mid y)$, with factorized transitions across positions.
Using Jensen's inequality with the forward conditional path law
$\qtildez_{1:T\mid 0,y}(x_{1:T}\mid x_0,y)$, we obtain
\begin{equation}
\label{eq:app_elbo_jensen_userstyle}
\begin{aligned}
\mE\big[-\log \pthetax_{0\mid y}(X_0\mid Y)\big]
&\le
-\mE\Bigg[
\log \frac{\pthetax_{\mid y}(X_{0:T}\mid Y)}
{\qtildez_{1:T\mid 0,y}(X_{1:T}\mid X_0,Y)}
\Bigg]
\\
&\leq
-\mE\Bigg[
\log \pthetax_{T\mid y}(X_T\mid Y)
+\sum_{t=1}^T
\log \frac{\pthetax_{t-1\mid t,y}(X_{t-1}\mid X_t,Y)}
{\qtildez_{t\mid t-1,y}(X_t\mid X_{t-1},Y)}
\Bigg]\eqsp.
\end{aligned}
\end{equation}
Using Bayes' rule,
\begin{equation}
\qtildez_{t\mid t-1,y}(x_t\mid x_{t-1},y)
=
\frac{\qtildez_{t-1\mid t,y}(x_{t-1}\mid x_t,y)\ \qtildez_{t\mid y}(x_t\mid y)}
{\qtildez_{t-1\mid y}(x_{t-1}\mid y)}\eqsp,
\end{equation}
and rearranging terms yields the standard ELBO decomposition:
\begin{equation}
\label{eq:app_elbo_kl_form_userstyle}
\begin{aligned}
\mE\big[-\log \pthetax_{0\mid y}(X_0\mid Y)\big]
&\le
\mE\Big[\KL\big(\qtildez_{T\mid y}(\cdot\mid Y)\ \big|\ \pthetax_{T\mid y}(\cdot\mid Y)\big)\Big]
+
\mE\big[\entropy(\qtildez_{0\mid y}(\cdot\mid Y))\big]
\\
&\quad
+\sum_{t=1}^T
\mE\Big[\KL\big(\qtildez_{t-1\mid t,y}(\cdot\mid X_t,Y)\ \big|\ \pthetax_{t-1\mid t,y}(\cdot\mid X_t,Y)\big)\Big]\eqsp.
\end{aligned}
\end{equation}
and the ELBO loss $\lossx$ is the RHS of this expression. 
In masked diffusion with $\bar\gamma_T=0$, both $\qtildez_{T\mid y}(\cdot\mid y)$ and $\pthetax_{T\mid y}(\cdot\mid y)$ are Dirac at the mask state, for any $y \in \latspace$, hence the first KL term is $0$.

\paragraph{Step 3: factorization lower bound.}
Since $\pthetax_{t-1\mid t,y}(\cdot\mid X_t,Y)$ factorizes across positions, for each $(t,X_t,Y)$ we have
\begin{align}
\label{eq:app_factorized_kl_lb_userstyle}
\KL\big(\qtildez_{t-1\mid t,y}(\cdot\mid X_t,Y)\ \big|\ \pthetax_{t-1\mid t,y}(\cdot\mid X_t,Y)\big)
&\ge \KL\big(\qtildez_{t-1\mid t,y}(\cdot\mid X_t,Y)\ \big|\
\prod_{i=1}^S \qtildezi_{t-1\mid t,y}(\cdot\mid X_t,Y)\big)\eqsp,
\end{align}
where $\qtildezi_{t-1\mid t,y}$ denotes the $i$-th marginal of $\qtildez_{t-1\mid t,y}$.
Plugging \eqref{eq:app_factorized_kl_lb_userstyle} into \eqref{eq:app_elbo_kl_form_userstyle} yields
\begin{align}
\lossx(\theta)
&\ge
\mE\big[\entropy(\qtildez_{0\mid y}(\cdot\mid Y))\big]
+\sum_{t=1}^T
\mE\Big[
\KL(\qtildez(X_{t-1}\mid X_t,Y) \Big| \prod_{i=1}^S \qtildezi(X_{t-1}^i\mid X_t,Y))
\Big]
\end{align}
where $\qtildez = \qtildez_{t-1\mid t,y}$ and $\qtildezi = \qtildezi_{t-1\mid t,y}$ (subscripts omitted for space), which is exactly the bound in \Cref{prop:latent_factorization_limits}.
\end{proof}

\section{LADD general framework}
\label{app:ladd_section}

\subsection{Principled ELBO loss}
\label{app:ladd_elbo}
We justify the training objective used for LADD by deriving a variational bound for a two-channel diffusion process $Z=(X,Y)$, where $X$ lies in the token space $\sspace^S$ and $Y$ lies in an auxiliary latent space $\latspace$.
The latent space $\latspace$ may be continuous or discrete; the derivation below is agnostic to this choice.

\paragraph{Forward process.}
Let $X_0\sim\qx_0$ and let $\enc_\varphi(\cdot\mid X_0)$ be an encoder distribution on $\latspace$, from which we sample $Y_0\sim\enc_\varphi(\cdot\mid X_0)$.
Denote the induced joint distribution at time $0$ by
\begin{equation}
\label{eq:app_qz0_ladd}
\qz_0(x_0,y_0)=\qx_0(x_0)\,\enc_\varphi(y_0\mid x_0)\eqsp.
\end{equation}
We consider a forward noising process over $Z_t=(X_t,Y_t)$ with \emph{conditionally independent channels}:
\begin{equation}
\label{eq:app_forward_ladd}
\qz(z_{0:T})
=
\qz_0(z_0)\,
\qx_{|0}(x_{1:T}\mid x_0)\,
\qy_{|0}(y_{1:T}\mid y_0)\eqsp,
\end{equation}
where $\qx_{|0}$ is the masked forward diffusion on $\sspace$ (\eqref{eq:app_mdlm_forward}--\eqref{eq:app_mdlm_marginal}), and $\qy_{|0}$ is either a Gaussian forward diffusion (\eqref{eq:app_forward_step}--\eqref{eq:app_forward_marginal}) in the continuous case, or a masked forward diffusion on $\sspace^M$ in the discrete-latent case.

\paragraph{Reverse model.}
We define a joint Markov reverse model
\begin{equation}
\label{eq:app_reverse_ladd}
\ptheta(z_{0:T})=\ptheta_T(z_T)\,\prod_{t=1}^T \ptheta_{t-1\mid t}(z_{t-1}\mid z_t)\eqsp,
\end{equation}
where $\ptheta_T$ is the terminal initialization.
For the \emph{joint} strategy (main text \Cref{sec:ladds_framework}), we use transitions that factorize across channels given $z_t$:
\begin{align}
\label{eq:app_reverse_factorize_channels}
&\ptheta_{t-1\mid t}(x_{t-1},y_{t-1}\mid x_t,y_t)
\\ &=\pthetax_{t-1\mid t}(x_{t-1}\mid x_t,y_t)\,
\pthetay_{t-1\mid t}(y_{t-1}\mid y_t,x_t)\eqsp.
\end{align}
(Within-channel factorization across positions is handled at the parameterization level and is not needed for the ELBO identity.)

\paragraph{Variational bound.}
For any $x_0\in\sspace^S$, by marginalizing out latents and trajectories,
\begin{equation}
\label{eq:app_marginal_px0}
\pthetax_0(x_0)
=
\int \ptheta(z_{0:T})\,\dd z_{1:T}\,\dd y_0\eqsp,
\end{equation}
where $\dd y_0$ denotes integration or summation with respect to the reference measure on $\latspace$ (Lebesgue in the continuous case, counting in the discrete case).
Multiplying and dividing by the forward conditional path density $\qx_{|0}(x_{1:T}\mid x_0)\qy_{|0}(y_{1:T}\mid y_0)$ and by $\enc_\varphi(y_0\mid x_0)$, Jensen's inequality gives
\begin{align}
\label{eq:app_jensen_ladd}
-\log \pthetax_0(x_0)
&\le
-\mE\Bigg[
\log(\ptheta(Z_{0:T})) - \log(\qx_{|0}(X_{1:T}\mid x_0)) \\
&- \log(\qy_{|0}(Y_{1:T}\mid Y_0)) - \log(\enc_\varphi(Y_0\mid x_0))
\ \Bigg|\ X_0=x_0
\Bigg]\eqsp,
\end{align}
where the expectation is over $Y_0\sim\enc_\varphi(\cdot\mid x_0)$, $X_{1:T}\sim\qx_{|0}(\cdot\mid x_0)$, and $Y_{1:T}\sim\qy_{|0}(\cdot\mid Y_0)$.
Taking expectation over $X_0\sim\qx_0$ yields the standard negative ELBO.

\paragraph{ELBO decomposition and additivity across channels.}
Using the path decompositions of each channel (continuous: \eqref{eq:app_path_decomp}; discrete: \eqref{eq:app_mdlm_elbo_decomp}) and the factorized forward \eqref{eq:app_forward_ladd}, the negative ELBO decomposes as
\begin{equation}
\label{eq:app_elbo_decomp_ladd}
\begin{aligned}
\mE\big[-\log \pthetax_0(X_0)\big]
&\le
\underbrace{\mE\big[\log \enc_\varphi(Y_0\mid X_0)\big]}_{L_{\mathrm{enc}}}
\\
&\quad+
\underbrace{\lossx_T+\lossx_0+\sum_{t=2}^T \lossx_{t-1}}_{\text{token-channel ELBO}}
+
\underbrace{\lossy_T+\lossy_0+\sum_{t=2}^T \lossy_{t-1}}_{\text{latent-channel ELBO}}\eqsp.
\end{aligned}
\end{equation}
where $\lossx_T,\lossx_0,\lossx_{t-1}$ are the usual ELBO terms for masked discrete diffusion (with reverse kernels possibly conditioned on $Y_t$), and $\lossy_T,\lossy_0,\lossy_{t-1}$ are the usual ELBO terms for the latent diffusion (with reverse kernels possibly conditioned on $X_t$).
Crucially, \eqref{eq:app_elbo_decomp_ladd} shows that, up to the encoder term $L_{\mathrm{enc}}$ and endpoint terms, the ELBO is a \emph{sum of per-channel diffusion objectives}.
This is the formal justification for optimizing a weighted sum of a token diffusion loss and a latent diffusion loss.

\paragraph{Instantiations: Co-LADD and Di-LADD.}
The generic per-time KL terms $\lossx_{t-1}$ and $\lossy_{t-1}$ reduce to the familiar closed-form diffusion losses once we adopt the standard bridge-based parameterizations.

\emph{Token channel.}
With the bridge substitution parameterization \eqref{eq:app_mdlm_param} (now conditioned on the chosen variables, e.g.\ $(X_t,Y_t)$ in the joint strategy), we recover the weighted masked cross-entropy form
\begin{equation}
\label{eq:app_ladd_token_ce}
\lossx_{t-1}
=
\mE\Bigg[
\lambdaxelbo_t
\sum_{i=1}^S -\mathbf{1}\{X_t^i=\maskvec\}\,
\log\big\langle \xpred^i(C_t,t),\,X_0^i\big\rangle
\Bigg]\eqsp,
\end{equation}
where $C_t=(X_t, Y_t)$ or $C_t=(X_t, Y_0)$ depending on the strategy, and $\lambdaxelbo_t$ given by \eqref{eq:app_mdlm_lambda} (or any reweighting).

\emph{Latent channel: Co-LADD.}
If $Y$ is continuous and we use the Gaussian bridge parameterization \eqref{eq:app_reverse_param}, then $\lossy_{t-1}$ reduces (up to constants) to a weighted regression objective as in \eqref{eq:app_y0_pred_weight}:
\begin{equation}
\label{eq:app_ladd_latent_mse}
\lossy_{t-1}
=
\mE\Big[
\lambdayelbo_t\,\|Y_0-\ypred(C_t,t)\|^2
\Big]\ +\ \text{const}\eqsp,
\end{equation}
where $C_t = (X_t, Y_t)$ or $Y_t$ depending on the process strategy. 

\emph{Latent channel: Di-LADD.}
If $Y$ is discrete and we use the masked bridge substitution (the analogue of \Cref{eq:app_mdlm_param} on the latent alphabet), then $\lossy_{t-1}$ reduces to the same weighted masked cross-entropy form as \eqref{eq:app_ladd_token_ce}.

\paragraph{Sequential strategy.}
The sequential strategy is handled similarly by applying the same ELBO argument to the factorized generative model
$\ptheta(y_{0:T^{\latspace}})\,\ptheta(x_{0:T^{\sspace}}\mid y_0)$.
The resulting bound again decomposes into (i) an ELBO for the latent diffusion and (ii) an ELBO for the token diffusion conditioned on the sampled $Y_0$, yielding the same additive structure as \eqref{eq:app_elbo_decomp_ladd} (with different conditioning variables), and leading to the objectives reported in the main text.

\paragraph{Endpoint and encoder terms.}
In practice we drop endpoint terms ($\lossx_T,\lossx_0,\lossy_T,\lossy_0$) as is standard in diffusion training (\eqref{eq:ddpm_elbo_appendix}, \eqref{eq:app_mdlm_elbo_decomp}).
The encoder term $L_{\mathrm{enc}}=\mE[\log \enc_\varphi(Y_0\mid X_0)]$ depends on the encoder design; for deterministic discrete encoders (e.g.\ vector quantization), it is constant under the counting measure, while for fixed-variance Gaussian encoders it contributes only an additive constant.
These terms do not affect the validity of optimizing the sum of channel-wise diffusion losses.

\subsection{Vector quantization for Di-LADD}
\label{app:vq}

Di-LADD relies on a vector quantizer to map continuous latent representations to a discrete latent sequence $Y_0\in\{0,\ldots,C-1\}^M$ and corresponding embeddings in $\rset^{d_\ell}$.

Given encoder outputs $\hat W_0=(\hat W_0^1,\ldots,\hat W_0^M)\in(\rset^{d_\ell})^M$, vector quantization returns discrete codes
$Y_0=(Y_0^1,\ldots,Y_0^M)\in\{0,\ldots,C-1\}^M$ and embeddings
$W_0=(W_0^1,\ldots,W_0^M)\in(\rset^{d_\ell})^M$ with $W_0^j=E_{Y_0^j}$.

\paragraph{Quantization map.}
Let $\hat{W}_0=(\hat{W}_0^1,\ldots,\hat{W}_0^M)\in(\rset^{d_\ell})^M$ denote the pre-quantized latent sequence produced by the Di-LADD encoder network.
Let $E=(E_0,\ldots,E_{C-1})\in\rset^{C\times d_\ell}$ be the learnable codebook, where $E_c\in\rset^{d_\ell}$ is the embedding of code index $c$. 

For each latent position $j$, we compute the nearest code index
\begin{equation}
\label{eq:vq_assignment}
Y_0^j \in \arg\min_{c\in\{0,\ldots,C-1\}} d(\hat{W}_0^j, E_c)\eqsp,
\end{equation}
and define the quantized latent as $W_0^j = E_{Y_0^j}$.
In our implementation, we use cosine similarity to define the matching metric (equivalently, $d(\cdot,\cdot)$ is induced by cosine distance after $\ell_2$-normalization).

\paragraph{Straight-through estimator and commitment loss.}
We use the straight-through estimator, treating the quantization map as the identity in the backward pass.
Training includes a standard VQ regularizer that encourages encoder outputs to commit to selected codes.
In the notation above, this takes the form
\begin{equation}
\label{eq:vq_commitment}
\loss_{\mathrm{VQ}}(\varphi)=\|\hat{W}_0 - \mathrm{sg}[W_0]\|^2\eqsp,
\end{equation}
where $\mathrm{sg}[\cdot]$ denotes stop-gradient. We update the codebook $E$ via EMA statistics rather than gradient descent, hence we do not include an explicit codebook loss term.

\paragraph{Dead-code handling.}
We use dead-code revival, with hyper-parameters listed in \Cref{app:experiments}. 

\paragraph{Learnable mask embedding.}
To support masked diffusion on the discrete latent channel, we add a dedicated mask embedding vector for the latent-mask state.
Concretely, the Di-LADD denoiser consumes a learnable vector representation for masked latent positions (and analogously for masked tokens in the data channel), rather than a fixed embedding.

\subsection{Further training considerations}
\label{app:further_training_considerations}

This section collects training-time techniques and schedule choices that affect stability and the strength of latent conditioning in LADD. Unless explicitly stated otherwise, the main-text results use the default choices summarized in \Cref{app:experiments}.

\paragraph{Latent-path robustness.}
We use classifier-free training as a lightweight robustness regularizers, to stabilize conditioning under partial information. We use a classifier-free drop-out probability $p_{\mathrm{cfg}}$ at training \citep{ho2022classifierfreediffusionguidance}, but use no guidance at inference. Additionally, we apply stochastic input dropout to the encoder, controlled by a parameter $p_r$.

\paragraph{Independent timesteps.}
We consider an \emph{independent timesteps} strategy, as inspired by \citet{campbell2024generative}, where we draw $t_x\sim\omega_x$ and $t_y\sim\omega_y$ independently, and apply the corresponding reconstruction losses. This leaves the joint/sequential coupling and noise schedules as inference-time choices. 

For Di-LADD, this variant implicitly provides latent-unconditional training cases: when $t_y=T^{\latspace}$ the latent is fully masked ($Y_{T^{\latspace}}=\mask$ almost surely), so the token loss includes trajectories in which the latent carries no information. In this sense, Di-LADD can obtain a classifier-free effect without explicit latent dropout.

For Co-LADD, the analogous extreme-noise latent $Y_{T^{\latspace}}$ remains a continuous random vector rather than a distinguished all-mask symbol. We therefore rely on $p_{\mathrm{cfg}}$ for Co-LADD in the main experiments.

\paragraph{Noise schedules and time sampling.}
We use separate noise schedules for the token and latent channels. 
We use a \emph{linear} noise schedule on the data channel, a \emph{linear} noise schedule on the latent channel for Di-LADD and a \emph{sqrt} noise schedule on the latent channel for Co-LADD, so latents are denoised slightly faster. For continuous diffusion (Co-LADD latent channel), we write $\tau=t/T\in[0,1]$ and specify the variance-preserving (VP) schedule via $\bar\alpha_t^2$. For masked discrete diffusion (token channel, and Di-LADD latent channel), we specify the schedule via the survival probability $\bar\gamma_t\in[0,1]$, with $\bar\gamma_0=1$ and $\bar\gamma_T=0$. We list classical noise schedules in \Cref{tab:noise_schedules}. We use a low-discrepancy time sampler with uniform time distribution \citep{kingma2023variationaldiffusionmodels}.

\paragraph{Two-stage training.}
To avoid early-stage instabilities such as latent collapse, we adopt a two-stage schedule inspired by \citep{rombach2022highresolutionimagesynthesislatent,vahdat2021scorebasedgenerativemodelinglatent}. 
In Stage 1, we start with $\lambdalatent=0$ to prioritize data reconstruction and form a meaningful latent structure. Then, in Stage 2, we optimize the full objective \eqref{eq:ladds_practical_objective} starting from the Stage~1 checkpoint and freezing the encoder; we leave more elaborate schedules to future work.

\paragraph{Co-LADD: Normalization.}
We observed a latent blow-up failure mode in Co-LADD using the joint process in Stage 1: the encoder increases the latent magnitude to maintain a high signal-to-noise ratio throughout the forward process, gaming data loss. This makes Stage 2 too hard as the denoiser must resolve the latent-embedding semantics very early during generation. To prevent this, we normalize encoder outputs, ensuring:
\begin{equation}
\label{eq:ladds_latent_norm}
\|\mu_{\varphi}(x_0)_0^j\|_2=1\qquad\text{for all }j\in\{1,\ldots,M\}\eqsp.
\end{equation}

\paragraph{\texorpdfstring{$p_{\mathrm{cfg}}$ (latent dropout for Co-LADD).}{p\_cfg (latent dropout for Co-LADD).}}
In Co-LADD, the latent channel is continuous and produced by an encoder as $Y_0\in(\rset^{d_\ell})^M$. To ensure the token denoiser does not over-rely on latents, we train it to also model the latent-unconditional distribution. Concretely, independently for each training sequence, with probability $p_{\mathrm{cfg}}$ we replace the encoder output by a fixed null latent,
\begin{equation}
Y_0 \leftarrow 0\eqsp,
\end{equation}
and then run the standard forward noising and diffusion losses using this $Y_0$. This is the usual classifier-free training pattern adapted to cross-modal conditioning \citep{zhou2025coevolutionarycontinuousdiscretediffusion}. At inference time, conditional and unconditional predictions can be combined via standard classifier-free guidance, but we do not explore guidance in this work.

\paragraph{\texorpdfstring{$p_r$ (encoder input dropout).}{p\_r (encoder input dropout).}}
We additionally regularize the latent pathway by randomly removing local information from the encoder input at training time. For each training sequence, we sample a dropout rate $p_r\sim \mathcal{U}[0,0.9]$. We then run the token forward process to obtain $X_t$ and form the encoder input sequence by taking the token embeddings and blanking a random subset of positions that are masked in $X_t$: for each position $i$ such that $X_t^i=\maskvec$, we set its encoder input embedding to zero with probability $p_r$ (independently across masked positions). The encoder therefore receives a partially blanked version of the already-masked sequence, which encourages $Y_0$ to be stable when local evidence is missing and discourages it from encoding brittle token-level details.

When $M=S$ (token-aligned latents), we optionally apply the same blanking to the corresponding encoder hidden states before they are used by the diffusion backbone, preserving alignment between blanked token positions and blanked latent positions. When $M<S$ (pooled latents), this post-encoder blanking is not well-defined without alignment and we omit it. See \Cref{app:neural_network_architectures} for the encoder/pooling definitions and alignment assumptions.

\begin{table}[t]
\centering
\scalebox{0.8}{
\begin{tabular}{lcc}
\toprule
Schedule & Name & Closed form \\
\midrule
Continuous & VP-Linear & $\bar{\alpha}_t^2 = 1-\tau$ \\
Continuous & VP-Cosine & $\bar{\alpha}_t^2 = \cos^2\!\Big(\frac{\tau+s}{1+s}\frac{\pi}{2}\Big)$ \\
Continuous & VP-Sqrt & $\bar{\alpha}_t^2 = \sqrt{1-\tau}$ \\
\midrule
Discrete & Linear & $\bar{\gamma}_t = 1-\tau$ \\
Discrete & Polynomial & $\bar{\gamma}_t = 1-\tau^2$ \\
Discrete & Geometric & $\bar{\gamma}_t = \exp\!\big(-\beta_{\min}^{\,1-\tau}\,\beta_{\max}^{\,\tau}\big)$ \\
\bottomrule
\end{tabular}
}
\vspace{5pt}
\caption{Schedule choices used in this work. We write $\tau=t/T\in[0,1]$. For VP-Cosine we use $s=0.008$. For the geometric discrete schedule, $\beta_{\min},\beta_{\max}>0$ are tunable endpoints.}
\label{tab:noise_schedules}
\end{table}

\section{Neural network architectures}
\label{app:neural_network_architectures}

\paragraph{DiT (single modality).}
We use the Diffusion Transformer (DiT) blocks of \citet{peebles2023scalablediffusionmodelstransformers} with bidirectional self-attention, RoPE \citep{su2023roformerenhancedtransformerrotary} for positional encoding and sinusoidal embeddings for timestep conditioning through AdaLN-Zero. This backbone is used as the MDLM denoiser, and (in the sequential strategy) as the latent denoiser when we generate the latent channel independently before denoising the token channel.

\paragraph{MM-DiT (multi modality).}
To evolve data and latent channels jointly, we adopt the multi-modal DiT (MM-DiT) of \citet{esser2024scalingrectifiedflowtransformers}. The input is formed by concatenating the $S$ data positions and the $M$ latent positions into a single sequence of length $(S+M)$, and we apply full self-attention on this concatenated sequence. Following the MM-DiT design (and mirroring the official implementation), transformer blocks are \emph{modality-specific}: each modality has its own MLP blocks and its own attention heads (and associated projections), while attention is nonetheless computed over the full concatenated sequence so that tokens and latents can interact at every layer. Input embeddings, layer norms, and timestep conditioning are also modality-specific. Outputs are produced by two modality-specific heads: a token head yielding $\xpred$ for the data channel, and a latent head yielding the appropriate latent predictor for the latent channel (specified below).

\paragraph{Output heads and latent type}
We use different latent heads depending on whether the latent channel is continuous (Co-LADD) or discrete (Di-LADD), while the token head is shared across all variants.

For the token channel, the MM-DiT (or DiT in MDLM) outputs $\xpred(\cdot,t)\in\simplex_K^S$, interpreted as per-position categorical predictions on the token alphabet with the mask coordinate set to $0$ at each position (SUBS-parameterization \citep{sahoo2024simpleeffectivemaskeddiffusion}), as in \Cref{sec:mdlm}.

For Co-LADD, the latent channel is a continuous sequence $Y\in(\rset^{d_\ell})^M$. The latent head outputs $\ypred(\cdot,t)\in(\rset^{d_\ell})^M$, interpreted as a $y_0$-prediction target used to parameterize the Gaussian reverse kernel (cf.\ \Cref{sec:ddpm} and \Cref{sec:ladds_objectives}).

For Di-LADD, the latent channel is a discrete sequence of code indices $Y\in\{0,\ldots,C-1\}^M$ (with a dedicated latent-mask state). The latent head outputs $\ypred(\cdot,t)\in\simplex_C^M$, interpreted as per-position categorical predictions on code indices with the latent-mask coordinate set to $0$ at each position, and used to parameterize the masked discrete reverse kernel on latents (cf.\ \Cref{sec:mdlm} and \Cref{sec:ladds_objectives}).

\paragraph{Joint vs.\ sequential usage.}
In the joint strategy and for sequential Di-LADD, we use a single MM-DiT to parameterize both reverse kernels, with cross-channel interaction enabled by full attention over the concatenated sequence. For sequential Co-LADD , we first denoise the latent channel with a latent-only DiT or MLP network, producing a clean latent sample, and we then denoise the token channel with an MM-DiT that conditions on this clean latent, implemented by including the latent positions in the concatenated sequence and holding them fixed across the token denoising trajectory.

\paragraph{Learned encoder.}
When training end-to-end, we implement the encoder $\enc_{\varphi}(\cdot\mid X_0)$ with a DiT-style network that consumes the $S$ input tokens. The encoder output is read from the final hidden states as the $M$ rightmost positions, yielding a latent representation in $(\rset^{d_\ell})^M$ after a linear projection to width $d_\ell$. 

In Di-LADD, the projected latent vectors are passed through vector quantization to obtain a discrete latent sequence (cf.\ \Cref{app:vq}).

\paragraph{Off-the-shelf encoder and pooling strategy.}
For text experiments, we replace the learned encoder with a frozen pretrained sentence encoder (Qwen3-Embedding 0.6B; \citep{qwen3technicalreport}) to obtain latent representations without end-to-end representation learning. Let the encoder produce token-level hidden states at the last layer. When $M=S$, we keep token-level alignment and take the last-layer hidden state at each of the $S$ token positions, clip each vector to its first $d_\ell$ dimensions, and apply per-token $\ell_2$ normalization. When $M<S$, we use an average pooling strategy, which down-samples the token-level hidden states into $M$ pooled vectors by averaging over adaptively sized contiguous chunks, while preserving the final EOS embedding as the last pooled element. This yields a latent sequence in $(\rset^{d_\ell})^M$ that can be consumed by Co-LADD directly or quantized for Di-LADD.

\section{Experimental setup}

\label{app:experiments}

All models use the architectures described in \Cref{app:neural_network_architectures}. Unless stated otherwise, we provide timestep conditioning as the normalized scalar $t/T$ (and the corresponding sinusoidal embeddings used by the backbone). We employ a common design choice in masked discrete diffusion by removing explicit timestep conditioning from the token denoiser $\xpred$ \citep{zheng2025masked}. 

\paragraph{Training horizons, schedules, and defaults.}
We use a training horizon of $T=4000$ for both channels. Sampling horizons and the inference budgets reported in the main text follow \Cref{sec:exp} and are not repeated here. For Di-LADD we use a linear noise schedule on the token channel and a linear noise schedule on the discrete latent channel. For Co-LADD we use a linear noise schedule on the token channel and a square-root schedule on the continuous latent channel (definitions and variants are given in \Cref{app:further_training_considerations}). For the continuous latent diffusion objective we use $y_0$-prediction. We use constant reconstruction weights $\lambda_t^x=1$ and $\lambda_t^y=1$ as the default in all main-text results. We use $p_{\text{cfg}} = 0.15$ and the $p_r$ corruption mechanism following \citep{zhou2025coevolutionarycontinuousdiscretediffusion}, though we do not observe noticeable improvements with or without them. 

\paragraph{Precision and entropy stability.}
Training is performed in \texttt{bf16} precision with gradient clipping at $1.0$. Sampling is performed in high precision \texttt{fp64} arithmetic. In our setup this does not induce entropic collapse; see \Cref{tab:lm1b-entropy-matched-flops}. 

\paragraph{Vector quantizer}
We implement vector quantization using the \texttt{VectorQuantize} module from the \texttt{vector-quantize-pytorch} library by \citet{wang2020vectorquantize}, and use the straight-through estimator to propagate gradients through the discrete code assignments.
We use EMA codebook updates with decay rate $0.95$, and a commitment cost $\beta=0.1$. Across experiments, we also vary $C\in\{512,2048\}$ and $d_\ell\in\{32,128\}$.
We also enable dead-code replacement with \texttt{threshold\_ema\_dead\_code}=2, which restarts rarely used codes according to the library's default heuristic.

\paragraph{FLOPs accounting.}
We report approximate forward-pass FLOPs using standard Transformer computations to reflect sole backbone cost, with the convention that one fused multiply-add counts as two FLOPs. For a sequence length $L$, hidden width $d$, and $n_{\ell}$ Transformer blocks, we approximate the cost of a block as the sum of (i) dense projections for attention ($Q,K,V$ and the output projection), which scale like $4Ld^2$, (ii) the attention mixing itself (forming $QK^{\top}$ and multiplying by $V$), which scales like $2L^2d$, and (iii) the MLP with expansion ratio $r$ (we take $r=4$), which scales like $2rLd^2$. This yields
\begin{equation}
    \mathrm{FLOPs}
\;\approx\;
2\,n_{\ell}\Big((4+2r)Ld^2 + 2L^2d\Big).
\end{equation}
We ignore LayerNorm, elementwise activations, bias additions, timestep/conditioning MLPs, and the input embedding/output vocabulary projection layers. 
This expression is essentially independent of the number of attention heads (for fixed $d$), since multi-head attention mostly redistributes the same computation across heads. For MM-DiT, we apply the same formula with the appropriate effective sequence length for the stream being processed (or the concatenation length when streams are combined).

\subsection{Low-dimensional dataset}
\label{app:exp_lowdim}

See \Cref{sec:lowdim_exp} for the definition of the data distribution $\qx_0$. The probability $\omega(i,y)$ of observing a $1$ at sequence position $i$ for a given latent shift $y$ is defined by
\begin{align}
\label{eq:omega_sawtooth}
&\omega(i,y)=s+(1-2s) \\
&\cdot\left[1-\mathrm{abs}\!\left(2\left(P\cdot\left(\frac{i-1}{S}+y\right)\bmod 1\right)-1\right)\right]\eqsp,
\end{align}
where $\mathrm{abs}$ is the absolute value, $i\in\{1,\ldots,S\}$, $y\in[0,1]$, $P$ is the number of sawtooth periods (we use $P=2$), and $s\in[0,0.5]$ is the minimum probability (we use $s=0.01$). We generate an effectively infinite stream of data by resampling at the end of each epoch.

\begin{figure}
    \centering
    \includegraphics[width=0.5\linewidth]{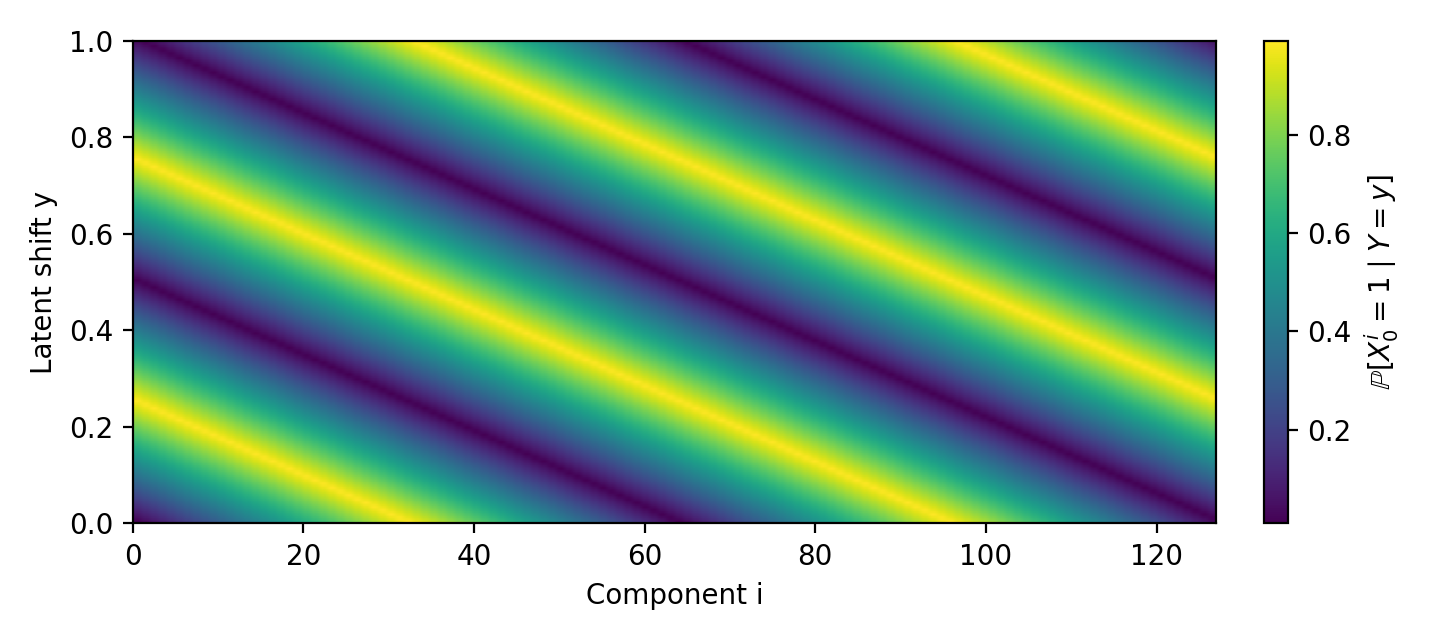}
    \caption{Values of $\qxi_{0|y}(1 | y)$ across coordinates for $y \in [0, 1]$.}
    \label{fig:sawtooth_distribution}
\end{figure}

We use sequence length $S=128$. The latent sequence length is $M=1$ with width $d_\ell=32$. We use codebook size $C = 2048$. 

For the backbones, we use MM-DiT (for joint models and for the token denoiser in the sequential strategy) and DiT (for the latent denoiser in the sequential strategy), with hidden size $64$, $4$ transformer layers, $4$ attention heads, and dropout $0.1$.

For sequential Co-LADD, we use a small MLP latent denoiser whose output conditions an MM-DiT token denoiser. We set $\lambda_t^x=\lambdaxelbo_t$ and $\lambda_t^y=1$.

\paragraph{Two-stage training.}
We use the two-stage procedure described in \Cref{sec:ladds_training}. Stage~1 runs for $20{,}000$ steps, and Stage~2 runs for $20{,}000$ steps, starting from the Stage~1 checkpoint. We use a $500$-step linear warmup at the start of training. We train with batch size $2048$ using AdamW (learning rate $1\!\times\!10^{-3}$, weight decay $0.01$); we maintain an exponential moving average (EMA) of model weights with decay $0.99$.

\paragraph{Evaluation.}
For evaluation, we generate $20{,}000$ samples and compute SWD using $1{,}000$ random Gaussian projection directions, as defined in \Cref{app:metrics}.

\begin{table}[t]
\centering
\caption{
LM1B validation PPL$\downarrow$ including both controlled Qwen3-tokenizer results and external GPT-2-tokenizer references. Results marked with ${}^*$ use the GPT-2 tokenizer. External references are included for context only: they differ from our controlled setting in tokenizer, vocabulary size, parameter count, training budget. ``elbo'' denotes training with $\lambda_t^x=\lambdaxelbo_t$; otherwise we use $\lambda_t^x=1$.
}
\label{tab:lm1b_validation_all}
\scalebox{0.8}{
\begin{tabular}{lcccc}
\toprule
Method & Tokenizer & Params & Train tokens & Validation PPL \\
\midrule
\multicolumn{5}{l}{\emph{External references, GPT-2 tokenizer}} \\
\midrule
AR$^*$ \citep{sahoo2024simpleeffectivemaskeddiffusion}
& GPT-2 & 92M & 33B & 22.32 \\
D3PM$^*$ \citep{austin2021structured}
& GPT-2 & 92M & 33B & $\leq 77.50$ \\
Diffusion-LM$^*$ \citep{li2022diffusionlmimprovescontrollabletext}
& GPT-2 & 92M & 33B & $\leq 118.62$ \\
DiffusionBert$^*$ \citep{he2022diffusionbertimprovinggenerativemasked}
& GPT-2 & 92M & 33B & $\leq 63.78$ \\
SEDD Absorb$^*$ \citep{lou2024discrete}
& GPT-2 & 92M & 33B & $\leq 32.79$ \\
MDLM$^*$ \citep{sahoo2024simpleeffectivemaskeddiffusion}
& GPT-2 & 92M & 33B & $\leq 27.03$ \\
\midrule
\multicolumn{5}{l}{\emph{Qwen3 tokenizer}} \\
\midrule
AR
& Qwen3 & 184M & 10B & 23.25 \\
\midrule
MDLM, elbo
& Qwen3 & 184M & 10B & $\leq 32.89$ \\
MDLM
& Qwen3 & 184M & 10B & $\leq 31.47$ \\
\midrule
Co-LADD-128, elbo
& Qwen3 & 184M & 10B & $\leq 28.88$ \\
Co-LADD-128
& Qwen3 & 184M & 10B & $\leq 23.66$ \\
Di-LADD-128
& Qwen3 & 184M & 10B & $\leq 12.28$ \\
Di-LADD-128, SEQ
& Qwen3 & 184M & 10B & $\leq 12.08$ \\
Co-LADD-32
& Qwen3 & 184M & 10B & $\leq 27.29$ \\
Di-LADD-32
& Qwen3 & 184M & 10B & $\leq 25.29$ \\
Di-LADD-32, SEQ
& Qwen3 & 184M & 10B & $\leq 24.71$ \\
\bottomrule
\end{tabular}
}
\end{table}

\subsection{Text dataset (LM1B)}
\label{app:exp_lm1b}

\begin{table*}[ht!]
\centering
\caption{
LM1B sampling metrics across inference-step budgets. We report Gen PPL (entropy below) and Throughput / Gen PPL 
(wall-clock time per sample below)
Best value is \textbf{bolded}, second best value is \underline{underlined}, and third best value is \emph{emphasized} among non-SEQ methods only; SEQ methods are shown separately and are left unranked.}
\label{tab:lm1b_gen_ppl_entropy_steps_time_results_1}
\resizebox{\textwidth}{!}{
\begin{tabular}{lrrrrrrrrrrrrrr}
\toprule
Method & \multicolumn{2}{c}{2} & \multicolumn{2}{c}{4} & \multicolumn{2}{c}{8} & \multicolumn{2}{c}{16} & \multicolumn{2}{c}{32} & \multicolumn{2}{c}{64} & \multicolumn{2}{c}{128} \\
\cmidrule(lr){2-3}\cmidrule(lr){4-5}\cmidrule(lr){6-7}\cmidrule(lr){8-9}\cmidrule(lr){10-11}\cmidrule(lr){12-13}\cmidrule(lr){14-15}
 & Gen PPL & Thr/Gen & Gen PPL & Thr/Gen & Gen PPL & Thr/Gen & Gen PPL & Thr/Gen & Gen PPL & Thr/Gen & Gen PPL & Thr/Gen & Gen PPL & Thr/Gen \\
\midrule
MDLM & \begin{tabular}[t]{@{}r@{}}\underline{3046.40} \\ {\scriptsize\color{gray}\itshape 4.439}\end{tabular} & \begin{tabular}[t]{@{}r@{}}\emph{0.3740} \\ {\scriptsize\color{gray} 0.878 ms}\end{tabular} & \begin{tabular}[t]{@{}r@{}}\textbf{1149.43} \\ {\scriptsize\color{gray}\itshape 4.424}\end{tabular} & \begin{tabular}[t]{@{}r@{}}\emph{0.4956} \\ {\scriptsize\color{gray} 1.755 ms}\end{tabular} & \begin{tabular}[t]{@{}r@{}}\textbf{462.47} \\ {\scriptsize\color{gray}\itshape 4.422}\end{tabular} & \begin{tabular}[t]{@{}r@{}}\emph{0.6159} \\ {\scriptsize\color{gray} 3.511 ms}\end{tabular} & \begin{tabular}[t]{@{}r@{}}\emph{295.55} \\ {\scriptsize\color{gray}\itshape 4.429}\end{tabular} & \begin{tabular}[t]{@{}r@{}}0.4819 \\ {\scriptsize\color{gray} 7.022 ms}\end{tabular} & \begin{tabular}[t]{@{}r@{}}\textbf{196.56} \\ {\scriptsize\color{gray}\itshape 4.387}\end{tabular} & \begin{tabular}[t]{@{}r@{}}\textbf{0.3623} \\ {\scriptsize\color{gray} 14.043 ms}\end{tabular} & \begin{tabular}[t]{@{}r@{}}\textbf{164.20} \\ {\scriptsize\color{gray}\itshape 4.370}\end{tabular} & \begin{tabular}[t]{@{}r@{}}\underline{0.2168} \\ {\scriptsize\color{gray} 28.087 ms}\end{tabular} & \begin{tabular}[t]{@{}r@{}}\textbf{154.91} \\ {\scriptsize\color{gray}\itshape 4.406}\end{tabular} & \begin{tabular}[t]{@{}r@{}}\underline{0.1149} \\ {\scriptsize\color{gray} 56.174 ms}\end{tabular} \\
Co-LADD-128 & \begin{tabular}[t]{@{}r@{}}\emph{3151.72} \\ {\scriptsize\color{gray}\itshape 4.450}\end{tabular} & \begin{tabular}[t]{@{}r@{}}0.3303 \\ {\scriptsize\color{gray} 0.961 ms}\end{tabular} & \begin{tabular}[t]{@{}r@{}}1205.49 \\ {\scriptsize\color{gray}\itshape 4.422}\end{tabular} & \begin{tabular}[t]{@{}r@{}}0.4317 \\ {\scriptsize\color{gray} 1.921 ms}\end{tabular} & \begin{tabular}[t]{@{}r@{}}532.32 \\ {\scriptsize\color{gray}\itshape 4.423}\end{tabular} & \begin{tabular}[t]{@{}r@{}}0.4888 \\ {\scriptsize\color{gray} 3.843 ms}\end{tabular} & \begin{tabular}[t]{@{}r@{}}331.38 \\ {\scriptsize\color{gray}\itshape 4.426}\end{tabular} & \begin{tabular}[t]{@{}r@{}}0.3926 \\ {\scriptsize\color{gray} 7.686 ms}\end{tabular} & \begin{tabular}[t]{@{}r@{}}253.67 \\ {\scriptsize\color{gray}\itshape 4.414}\end{tabular} & \begin{tabular}[t]{@{}r@{}}0.2565 \\ {\scriptsize\color{gray} 15.371 ms}\end{tabular} & \begin{tabular}[t]{@{}r@{}}197.59 \\ {\scriptsize\color{gray}\itshape 4.384}\end{tabular} & \begin{tabular}[t]{@{}r@{}}0.1646 \\ {\scriptsize\color{gray} 30.743 ms}\end{tabular} & \begin{tabular}[t]{@{}r@{}}198.79 \\ {\scriptsize\color{gray}\itshape 4.402}\end{tabular} & \begin{tabular}[t]{@{}r@{}}0.0818 \\ {\scriptsize\color{gray} 61.486 ms}\end{tabular} \\
Di-LADD-128 & \begin{tabular}[t]{@{}r@{}}\textbf{3004.26} \\ {\scriptsize\color{gray}\itshape 4.457}\end{tabular} & \begin{tabular}[t]{@{}r@{}}0.3723 \\ {\scriptsize\color{gray} 0.894 ms}\end{tabular} & \begin{tabular}[t]{@{}r@{}}\underline{1154.71} \\ {\scriptsize\color{gray}\itshape 4.431}\end{tabular} & \begin{tabular}[t]{@{}r@{}}0.4843 \\ {\scriptsize\color{gray} 1.788 ms}\end{tabular} & \begin{tabular}[t]{@{}r@{}}\underline{485.95} \\ {\scriptsize\color{gray}\itshape 4.400}\end{tabular} & \begin{tabular}[t]{@{}r@{}}0.5753 \\ {\scriptsize\color{gray} 3.577 ms}\end{tabular} & \begin{tabular}[t]{@{}r@{}}\underline{287.02} \\ {\scriptsize\color{gray}\itshape 4.404}\end{tabular} & \begin{tabular}[t]{@{}r@{}}\emph{0.4870} \\ {\scriptsize\color{gray} 7.153 ms}\end{tabular} & \begin{tabular}[t]{@{}r@{}}\underline{223.51} \\ {\scriptsize\color{gray}\itshape 4.406}\end{tabular} & \begin{tabular}[t]{@{}r@{}}0.3127 \\ {\scriptsize\color{gray} 14.307 ms}\end{tabular} & \begin{tabular}[t]{@{}r@{}}\underline{193.29} \\ {\scriptsize\color{gray}\itshape 4.394}\end{tabular} & \begin{tabular}[t]{@{}r@{}}0.1808 \\ {\scriptsize\color{gray} 28.613 ms}\end{tabular} & \begin{tabular}[t]{@{}r@{}}\underline{177.86} \\ {\scriptsize\color{gray}\itshape 4.376}\end{tabular} & \begin{tabular}[t]{@{}r@{}}0.0982 \\ {\scriptsize\color{gray} 57.227 ms}\end{tabular} \\
Co-LADD-32 & \begin{tabular}[t]{@{}r@{}}3313.72 \\ {\scriptsize\color{gray}\itshape 4.450}\end{tabular} & \begin{tabular}[t]{@{}r@{}}\textbf{0.4592} \\ {\scriptsize\color{gray} 0.657 ms}\end{tabular} & \begin{tabular}[t]{@{}r@{}}1238.70 \\ {\scriptsize\color{gray}\itshape 4.419}\end{tabular} & \begin{tabular}[t]{@{}r@{}}\textbf{0.6142} \\ {\scriptsize\color{gray} 1.314 ms}\end{tabular} & \begin{tabular}[t]{@{}r@{}}541.28 \\ {\scriptsize\color{gray}\itshape 4.417}\end{tabular} & \begin{tabular}[t]{@{}r@{}}\textbf{0.7028} \\ {\scriptsize\color{gray} 2.629 ms}\end{tabular} & \begin{tabular}[t]{@{}r@{}}337.95 \\ {\scriptsize\color{gray}\itshape 4.425}\end{tabular} & \begin{tabular}[t]{@{}r@{}}\underline{0.5629} \\ {\scriptsize\color{gray} 5.257 ms}\end{tabular} & \begin{tabular}[t]{@{}r@{}}292.62 \\ {\scriptsize\color{gray}\itshape 4.434}\end{tabular} & \begin{tabular}[t]{@{}r@{}}\emph{0.3250} \\ {\scriptsize\color{gray} 10.514 ms}\end{tabular} & \begin{tabular}[t]{@{}r@{}}225.40 \\ {\scriptsize\color{gray}\itshape 4.389}\end{tabular} & \begin{tabular}[t]{@{}r@{}}\emph{0.2110} \\ {\scriptsize\color{gray} 21.029 ms}\end{tabular} & \begin{tabular}[t]{@{}r@{}}216.27 \\ {\scriptsize\color{gray}\itshape 4.409}\end{tabular} & \begin{tabular}[t]{@{}r@{}}\emph{0.1099} \\ {\scriptsize\color{gray} 42.057 ms}\end{tabular} \\
Di-LADD-32 & \begin{tabular}[t]{@{}r@{}}3269.65 \\ {\scriptsize\color{gray}\itshape 4.457}\end{tabular} & \begin{tabular}[t]{@{}r@{}}\underline{0.4302} \\ {\scriptsize\color{gray} 0.711 ms}\end{tabular} & \begin{tabular}[t]{@{}r@{}}\emph{1193.28} \\ {\scriptsize\color{gray}\itshape 4.437}\end{tabular} & \begin{tabular}[t]{@{}r@{}}\underline{0.5894} \\ {\scriptsize\color{gray} 1.422 ms}\end{tabular} & \begin{tabular}[t]{@{}r@{}}\emph{503.86} \\ {\scriptsize\color{gray}\itshape 4.417}\end{tabular} & \begin{tabular}[t]{@{}r@{}}\underline{0.6979} \\ {\scriptsize\color{gray} 2.844 ms}\end{tabular} & \begin{tabular}[t]{@{}r@{}}\textbf{286.83} \\ {\scriptsize\color{gray}\itshape 4.398}\end{tabular} & \begin{tabular}[t]{@{}r@{}}\textbf{0.6130} \\ {\scriptsize\color{gray} 5.687 ms}\end{tabular} & \begin{tabular}[t]{@{}r@{}}\emph{248.06} \\ {\scriptsize\color{gray}\itshape 4.403}\end{tabular} & \begin{tabular}[t]{@{}r@{}}\underline{0.3544} \\ {\scriptsize\color{gray} 11.374 ms}\end{tabular} & \begin{tabular}[t]{@{}r@{}}\emph{194.62} \\ {\scriptsize\color{gray}\itshape 4.415}\end{tabular} & \begin{tabular}[t]{@{}r@{}}\textbf{0.2259} \\ {\scriptsize\color{gray} 22.749 ms}\end{tabular} & \begin{tabular}[t]{@{}r@{}}\emph{185.24} \\ {\scriptsize\color{gray}\itshape 4.403}\end{tabular} & \begin{tabular}[t]{@{}r@{}}\textbf{0.1187} \\ {\scriptsize\color{gray} 45.498 ms}\end{tabular} \\
\midrule
Di-LADD-128 (SEQ) & \begin{tabular}[t]{@{}r@{}}653.89 \\ {\scriptsize\color{gray}\itshape 4.433}\end{tabular} & \begin{tabular}[t]{@{}r@{}}0.7601 \\ {\scriptsize\color{gray} 2.012 ms}\end{tabular} & \begin{tabular}[t]{@{}r@{}}437.62 \\ {\scriptsize\color{gray}\itshape 4.443}\end{tabular} & \begin{tabular}[t]{@{}r@{}}0.7863 \\ {\scriptsize\color{gray} 2.906 ms}\end{tabular} & \begin{tabular}[t]{@{}r@{}}358.83 \\ {\scriptsize\color{gray}\itshape 4.436}\end{tabular} & \begin{tabular}[t]{@{}r@{}}0.5937 \\ {\scriptsize\color{gray} 4.694 ms}\end{tabular} & \begin{tabular}[t]{@{}r@{}}340.01 \\ {\scriptsize\color{gray}\itshape 4.440}\end{tabular} & \begin{tabular}[t]{@{}r@{}}0.3556 \\ {\scriptsize\color{gray} 8.271 ms}\end{tabular} & \begin{tabular}[t]{@{}r@{}}316.07 \\ {\scriptsize\color{gray}\itshape 4.436}\end{tabular} & \begin{tabular}[t]{@{}r@{}}0.2051 \\ {\scriptsize\color{gray} 15.424 ms}\end{tabular} & \begin{tabular}[t]{@{}r@{}}317.34 \\ {\scriptsize\color{gray}\itshape 4.434}\end{tabular} & \begin{tabular}[t]{@{}r@{}}0.1060 \\ {\scriptsize\color{gray} 29.731 ms}\end{tabular} & \begin{tabular}[t]{@{}r@{}}293.92 \\ {\scriptsize\color{gray}\itshape 4.424}\end{tabular} & \begin{tabular}[t]{@{}r@{}}0.0583 \\ {\scriptsize\color{gray} 58.345 ms}\end{tabular} \\
Di-LADD-32 (SEQ) & \begin{tabular}[t]{@{}r@{}}1606.59 \\ {\scriptsize\color{gray}\itshape 4.422}\end{tabular} & \begin{tabular}[t]{@{}r@{}}0.3891 \\ {\scriptsize\color{gray} 1.600 ms}\end{tabular} & \begin{tabular}[t]{@{}r@{}}681.43 \\ {\scriptsize\color{gray}\itshape 4.418}\end{tabular} & \begin{tabular}[t]{@{}r@{}}0.6352 \\ {\scriptsize\color{gray} 2.310 ms}\end{tabular} & \begin{tabular}[t]{@{}r@{}}363.85 \\ {\scriptsize\color{gray}\itshape 4.403}\end{tabular} & \begin{tabular}[t]{@{}r@{}}0.7364 \\ {\scriptsize\color{gray} 3.732 ms}\end{tabular} & \begin{tabular}[t]{@{}r@{}}253.81 \\ {\scriptsize\color{gray}\itshape 4.406}\end{tabular} & \begin{tabular}[t]{@{}r@{}}0.5991 \\ {\scriptsize\color{gray} 6.576 ms}\end{tabular} & \begin{tabular}[t]{@{}r@{}}227.83 \\ {\scriptsize\color{gray}\itshape 4.424}\end{tabular} & \begin{tabular}[t]{@{}r@{}}0.3579 \\ {\scriptsize\color{gray} 12.263 ms}\end{tabular} & \begin{tabular}[t]{@{}r@{}}215.63 \\ {\scriptsize\color{gray}\itshape 4.399}\end{tabular} & \begin{tabular}[t]{@{}r@{}}0.1962 \\ {\scriptsize\color{gray} 23.638 ms}\end{tabular} & \begin{tabular}[t]{@{}r@{}}193.73 \\ {\scriptsize\color{gray}\itshape 4.403}\end{tabular} & \begin{tabular}[t]{@{}r@{}}0.1113 \\ {\scriptsize\color{gray} 46.387 ms}\end{tabular} \\
\bottomrule
\end{tabular}
}
\end{table*}

We evaluate on LM1B \citep{chelba2014billionwordbenchmarkmeasuring}. Tokenization uses the Qwen3 tokenizer from HuggingFace; the vocabulary size is $K=151{,}670$ and includes the added \textsc{[mask]} token. We use packed sequences of length $S=128$.
Constant token-loss weighting, $\lambdax_t=1$, improves over the ELBO wieghting $\lambdaxelbo_t$ for both MDLM and Co-LADD (\Cref{tab:lm1b_validation_all}), and we use constant weighting with the remaining models.

\Cref{tab:lm1b_gen_ppl_entropy_steps_time_results_1} reports raw 
inference-time measurements for each method at each step budget, alongside 
Gen PPL and entropy. We use Throughput per Gen PPL point, defined as 
$1000 / (\mathrm{GenPPL} \cdot \mathrm{TotalMS})$, as a quality--speed metric: 
higher values indicate better text quality per unit of sampling time. The 
trends mirror the FLOPs-based results in \Cref{tab:lm1b-gen-ppl-matched-flops}: 
LADD variants dominate the few-step quality--speed frontier, with 
compressed-latent models offering the best per-time quality. SEQ variants 
are reported separately because they fix the latent denoising budget and 
vary only the token-step count.

\paragraph{Encoders and latent shapes.}
For Co-LADD and Di-LADD at scale, we use Qwen3-Embedding 0.6B \citep{qwen3technicalreport} as a frozen encoder to obtain latent representations. Since the embedding model is Matryoshka-style (MRT), we truncate the representation to width $d_\ell=32$ and apply $\ell_2$ normalization. As observed on the sawtooth experiment, we expect that a tailored encoder could improve performance further by making the conditional posterior more factorized; in this sense, the Qwen3 encoder is not necessarily optimal. We leave a systematic study of encoder design to future work. We use either (i) token-aligned latents with $M=S$ and codebook size $C=512$, or (ii) average pooled latents with $M=32<S$ and codebook size $C=2048$. Pooling strategy are further discussed in \Cref{app:neural_network_architectures}. 

\paragraph{Backbones and model variants.}
Joint LADD uses an MM-DiT joint denoiser over the concatenated $(S+M)$ sequence. Sequential LADD first denoises the latent channel with a latent-only DiT denoiser and then denoises the token channel with an MM-DiT using the resulting clean latent sample. MDLM uses a single-modality DiT denoiser. See \Cref{tab:model_sizes_and_configs} for full architectural hyperparameters (depth/width/heads/dropout etc.). 

\begin{table}[t]
\centering
\small
\setlength{\tabcolsep}{4pt}
\renewcommand{\arraystretch}{1.}
\caption{
Model configurations used in our text experiments. Parameter counts exclude input embedding layers and output projection head.
}
\resizebox{\linewidth}{!}{
\begin{tabular}{l l c c c c c c c}
\toprule
Method & Network(s) & $d$ & \#layers & \#heads & $d_{\mathrm{cond}}$ & dropout & $\text{mlp}_{\text{ratio}}$ & \#params \\
\midrule
MDLM &
Discrete DiT & 1024 & 13 & 16 & 256 & 0.1 & 4 & 184M \\
\midrule
Di-LADD / Co-LADD &
MM-DiT & 768 & 12 & 12 & 128 & 0.1 & 4 & 184M \\
\bottomrule
\end{tabular}
}
\label{tab:model_sizes_and_configs}
\end{table}

\paragraph{Training details.}
We train on $8$ H100 GPUs with per-GPU batch size $64$ (effective batch size $512$), for $300{,}000$ optimization steps (Stage 2 only) with effective batch size 512. This corresponds to a total of $10B$ tokens seen, under classical accounting rules \citep{sahoo2024simpleeffectivemaskeddiffusion}. We use the AdamW optimizer (learning rate $3\!\times\!10^{-4}$, weight decay $0.03$) and a $2{,}500$-step linear warmup followed by a constant learning rate. We set the random seed to $42$. We maintain an exponential moving average (EMA) of model weights with decay $0.9999$. We use \texttt{torch.compile}. Under these conditions, a single full run with no validation loop requires approximately $110$ H100-hours for LADD with MM-DiT and $M=128$, and $80$ H100-hours for MDLM with DiT.

\paragraph{Evaluation protocol.}
For evaluation, we sample $512$ sequences and vary the sampling budget in $\{1,2,4,8,16,32,64,128\}$ steps, as in the main text. Generative perplexity is measured under a fixed GPT-2 Large teacher \citep{radford2019language}. We also report token entropy computed from generated samples as detailed in \Cref{app:metrics}; in particular, \Cref{tab:lm1b-entropy-matched-flops} shows that our results do not exhibit entropic collapse under the reported settings, as entropy stays within 0.05 nats of MDLM at every matched-compute point. We also include per step sampling compute budget in \Cref{tab:gflops-per-step}.

\begin{table}[t]
\centering
\caption{
LM1B sample entropy at fixed sampling-compute budgets. Columns correspond to the same GFLOP budgets as in \Cref{tab:lm1b-gen-ppl-matched-flops}. Values are linearly interpolated between measured sampling budgets, and \texttt{n/a} is reported instead of extrapolation.
}
\label{tab:lm1b-entropy-matched-flops}
\scalebox{0.78}{
\begin{tabular}{lrrrrrrr}
\toprule
GFLOPs & 50 & 100 & 200 & 400 & 800 & 1600 & 2500 \\
\midrule
MDLM
& 4.436
& 4.424
& 4.423
& 4.422
& 4.384
& 4.376
& 4.400 \\
Co-LADD-128
& 4.448
& 4.422
& 4.423
& 4.425
& 4.411
& 4.386
& 4.397 \\
Di-LADD-128
& 4.455
& 4.428
& 4.401
& 4.404
& 4.405
& 4.393
& 4.381 \\
Di-LADD-128 (SEQ)
& \texttt{n/a}
& \texttt{n/a}
& 4.442
& 4.439
& 4.436
& 4.433
& 4.427 \\
Co-LADD-32
& 4.419
& 4.417
& 4.425
& 4.430
& 4.391
& \texttt{n/a}
& \texttt{n/a} \\
Di-LADD-32
& 4.442
& 4.422
& 4.402
& 4.402
& 4.413
& 4.405
& \texttt{n/a} \\
Di-LADD-32 (SEQ)
& 4.422
& 4.412
& 4.404
& 4.418
& 4.406
& 4.402
& \texttt{n/a} \\
\bottomrule
\end{tabular}
}
\end{table}

\begin{table}[t]
\centering
\caption{
Per-step sampling compute used for the matched-budget comparisons.
}
\label{tab:gflops-per-step}
\begin{minipage}{0.48\linewidth}
\centering
\scalebox{0.85}{
\begin{tabular}{lr}
\toprule
\multicolumn{2}{c}{LM1B} \\
\midrule
Method & GFLOPs / step \\
\midrule
MDLM & 21.37 \\
Co/Di-LADD-128 & 22.95 \\
Co/Di-LADD-32 & 14.06 \\
\bottomrule
\end{tabular}
}
\end{minipage}
\hfill
\begin{minipage}{0.48\linewidth}
\centering
\scalebox{0.85}{
\begin{tabular}{lr}
\toprule
\multicolumn{2}{c}{OWT} \\
\midrule
Method & GFLOPs / step \\
\midrule
MDLM & 90.73 \\
Co/Di-LADD-64 & 55.04 \\
Co/Di-LADD-16 & 49.98 \\
\bottomrule
\end{tabular}
}
\end{minipage}
\end{table}

\subsection{Text dataset (OWT)}
\label{app:exp_owt}

We evaluate on OpenWebText (OWT) \citep{Gokaslan2019OpenWeb} using the same 
setup as for LM1B (\Cref{app:exp_lm1b}), with the following changes: we use 
packed sequences of length $S=512$, latent length $M=16, 64$ for compressed 
latents, $500{,}000$ training steps, corresponding to $65.5$B tokens seen, and a sampling budget swept over 
$\{8, 32, 128, 512\}$ steps. The zero-shot perplexity evaluation uses the OWT checkpoints and evaluates them on OWT together with AG News \cite{zhang2016characterlevelconvolutionalnetworkstext}, PubMed \cite{cohan2018discourseawareattentionmodelabstractive}, Lambada \cite{paperno2016lambadadatasetwordprediction}, WikiText \cite{merity2016pointersentinelmixturemodels}, PTB \cite{marcus-etal-1993-building}. A single full run with no validation loop requires approximately $900$ H100-hours for LADD, and $650$ H100-hours for MDLM.

The tokenizer, encoder (frozen 
Qwen3-Embedding 0.6B), backbone configurations, optimizer settings, training 
hardware, and evaluation protocol (Gen PPL under GPT-2 Large, entropy 
measurement, sample size) are identical to LM1B.

\Cref{tab:owt-entropy-matched-flops} confirms that our OWT results do not 
exhibit entropic collapse: entropy remains within $0.1$ nats of MDLM at 
every matched-compute point, mirroring the LM1B finding.

\begin{table}[ht!]
\centering
\caption{
OWT sample entropy at fixed sampling-compute budgets. Columns correspond to the same GFLOP budgets as in \Cref{tab:owt-gen-ppl-matched-flops}. Values are linearly interpolated between measured sampling budgets, and \texttt{n/a} is reported instead of extrapolation.
}
\label{tab:owt-entropy-matched-flops}
\scalebox{0.8}{
\begin{tabular}{lrrrrrr}
\toprule
GFlops & 800 & 1600 & 3200 & 6400 & 12800 & 25600 \\
\midrule
MDLM & 5.49 & 5.44 & 5.35 & 5.33 & 5.28 & 5.28 \\
Co-LADD-16 & 5.45 & 5.35 & 5.34 & 5.33 & 5.31 & 5.28 \\
Di-LADD-16 & 5.46 & 5.40 & 5.36 & 5.30 & 5.30 & 5.28 \\
Di-LADD-16 (SEQ) & 5.46 & 5.42 & 5.41 & 5.38 & 5.37 & 5.34\\
Co-LADD-64 & 5.46 & 5.38 & 5.35 & 5.32 & 5.28 & 5.20 \\
Di-LADD-64 & 5.49 & 5.40 & 5.37 & 5.33 & 5.32 & 5.30 \\
Di-LADD-64 (SEQ) & 5.50 & 5.46 & 5.46 & 5.48 & 5.47 & 5.42 \\
\bottomrule
\end{tabular}
}
\end{table}

\subsection{Metrics}
\label{app:metrics}

We report standard metrics for discrete generative models, following common practice in masked discrete diffusion \citep{sahoo2024simpleeffectivemaskeddiffusion,zheng2025masked}. Throughout, a sequence is denoted $x_{1:S}\in\sspace^S$ and we omit special tokens (e.g.\ \textsc{[bos]}, \textsc{[eos]}, padding) whenever a metric is defined over 'content' tokens only.

\paragraph{Validation perplexity (ELBO-based).}
Given a validation set $\mathcal{D}=\{x_{1:S}^{(n)}\}_{n=1}^N$ and a trained diffusion model, we report a perplexity obtained from the data-channel variational bound. Concretely, let $\widehat{\mathrm{NLL}}_{\mathrm{ELBO},x}(x_{1:S})$ denote the estimated negative ELBO associated with the token channel (i.e., the discrete diffusion objective used for evaluation). We define
\begin{equation}
\label{eq:app_val_ppl}
\mathrm{PPL}_x(\mathcal{D})
=
\exp\left(
\frac{1}{NS}\sum_{n=1}^N
\widehat{\mathrm{NLL}}_{\mathrm{ELBO},x}\!\left(x_{1:S}^{(n)}\right)
\right)\eqsp.
\end{equation}
For LADD, $\widehat{\mathrm{NLL}}_{\mathrm{ELBO},x}$ corresponds to the token-channel bound $\lossx$ (and does not include the latent-channel ELBO term), so that $\mathrm{PPL}_x$ reflects token modeling quality under the diffusion objective.

\paragraph{Generative perplexity (teacher scoring).}
We also report a teacher-scored perplexity on model samples. Let $\widehat{\mathcal{X}}=\{x_{1:S}^{(n)}\}_{n=1}^N$ be a set of samples generated by the model and let $p_{\text{gen\_ppl}}$ be a fixed autoregressive teacher. We compute
\begin{align}
\label{eq:app_gen_ppl}
&\mathrm{GenPPL}(\widehat{\mathcal{X}})
\\&=
\exp\left(
\frac{1}{NS}\sum_{n=1}^N\sum_{s=1}^S
-\log p_{\text{gen\_ppl}}\!\left(x_s^{(n)}\mid x_{<s}^{(n)}\right)
\right)\eqsp.
\end{align}
When working with the Qwen3 tokenizer, we de-tokenize the generated sequences and re-tokenize with the GPT-2 tokenizer, and replace $S$ dynamically with the produced sequence length. In all text experiments, we use GPT-2 Large \citep{radford2019language} as the teacher $p_{\text{gen\_ppl}}$.

\paragraph{Token entropy (sample marginal).}
Teacher perplexity can be artificially improved by low-entropy sampling, so we also track the diversity of generated text via token entropy. From the same sample set $\widehat{\mathcal{X}}$, we form the empirical marginal distribution at each position $i$,
\begin{equation}
\label{eq:app_entropy_marginals}
\widehat{p}_i(k)
=
\frac{1}{N}\sum_{n=1}^N \mathbf{1}\!\{x_i^{(n)}=k\}\eqsp,
\quad k\in\sspace,\ \ i\in\{1,\ldots,S\}\eqsp,
\end{equation}
and we report the average positional entropy
\begin{equation}
\label{eq:app_entropy}
\mathrm{Ent}(\widehat{\mathcal{X}})
=
\frac{1}{S}\sum_{i=1}^S
\Big(-\sum_{k\in\sspace}\widehat{p}_i(k)\,\log \widehat{p}_i(k)\Big)\eqsp,
\end{equation}
with the convention $0\log 0=0$. \Cref{tab:lm1b-entropy-matched-flops} and \Cref{tab:owt-entropy-matched-flops} shows that our reported GenPPL improvements are not driven by entropic collapse.

\paragraph{Sliced Wasserstein distance (SWD).}
For $\mu,\nu\in\mathcal{P}(\rset^d)$ and $p\ge 1$, define
\begin{equation}
\mathbf{W}_p^p(\mu,\nu)=\inf_{\gamma\in\mathcal{M}(\mu,\nu)}\int \left\|x-y\right\|^p \gamma\!\left(\mathrm{d}x,\mathrm{d}y\right)\eqsp,
\end{equation}
and the sliced Wasserstein distance
\begin{equation}
\mathrm{SWD}_p(\mu,\nu)=\int_{\rset^{d}} \mathbf{W}_p\left(u_{\#}\mu,\  u_{\#}\nu\right) \normal(\dd u; 0, \Idd)\eqsp,
\end{equation}
where $u_{\#}\mu$ is the pushforward of $\mu$ by $x\mapsto \langle u,x\rangle$. In one dimension, for $\alpha,\beta\in\mathcal{P}(\rset)$,
\begin{equation}
\mathbf{W}_p^p(\alpha,\beta)=\int_0^1 \left|F_\alpha^{-1}(t)-F_\beta^{-1}(t)\right|^p \mathrm{d}t\eqsp,
\end{equation}
where $F^{-1}_{\alpha}, F^{-1}_{\beta}$ are the inverse CDFs of $\alpha, \beta$. Hence for empirical measures $\hat\alpha=\frac{1}{n}\sum_{i=1}^n \delta_{x_i}$ and $\hat\beta=\frac{1}{n}\sum_{i=1}^n \delta_{y_i}$,
\begin{equation}
\mathbf{W}_p^p(\hat\alpha,\hat\beta)=\frac{1}{n}\sum_{i=1}^n \left|x_{(i)}-y_{(i)}\right|^p\eqsp,
\end{equation}
with $x_{(i)},y_{(i)}$ the sorted samples. We report $\mathrm{SWD}_1$ estimated by Monte Carlo over directions $u\sim \normal(0,\Idd)$, averaging the corresponding one-dimensional $\mathbf{W}_1$ values.

\subsection{Additional results}

\label{app:additional_results}

\paragraph{Generated sequences on LM1B, uncurated.}

\textbf{MDLM}
\medbreak
\begin{tcblisting}{genbox}
<|endoftext|> home dive.<|endoftext|>The Kenyan Gesel, now 38, is living in a brand of a parallel life.<|endoftext|>They say water managers at the area's Becker River Management and Fish and Wildlife expect a hearing soon to decide that any fish endangered by runoff could be out of life.<|endoftext|>" The operation [Similar in 1990] was in good force.<|endoftext|>The council, chaired by Ban Ki-moon, said Saturday that Ethiopia had stepped up its commitment to al-Qaida and extremists as the "credible terror group" in the Horn of Africa.<|endoftext|>Under the agreement, CTCC already has Mad<|endoftext|>
\end{tcblisting}

\textbf{Co-LADD 128}
\medbreak
\begin{tcblisting}{genbox}
<|endoftext|> first it seems like the news always comes out.<|endoftext|>But that's how the White House has draft plans and hits a series of screen menacing signs of lawmakers complaining.<|endoftext|>The actual diagnosis of these issues doesn't carry over anytime soon, " said Sean Porteous, a U.S. government official.<|endoftext|>Lee will not play it from conquering Everest in Rome on July 89 in case of such a case that he is stuck in a likely hole in that path.<|endoftext|>A cameraman in Sanaa on Thursday ransacked the guest house of Lebanese prime minister Fouad Tweatt and arrested around 1,<|endoftext|>
\end{tcblisting}

\textbf{Co-LADD 32}
\medbreak
\begin{tcblisting}{genbox}
<|endoftext|> will be available.<|endoftext|>Kirughey (63) was enough to equalize their building as Tiger advances to the second round at the 15th hole when Steve Williams missed a controversial final birdie putt, although he got to settle for two bogies in his spectacular eagle at the par-4 16th.<|endoftext|>Okada birdied a 17th hole for the fourth helping Wales beat as Australia's Swans team beat England's twice winner Eddie Lewis 4-over 34, 10th in five three behind Perry and are three shots ahead of Chilean Brits Barnes.<|endoftext|><|endoftext|>
\end{tcblisting}

\textbf{Di-LADD 128}
\medbreak
\begin{tcblisting}{genbox}
<|endoftext|> of the White House in Washington -- a vast White House staff -- which are charged with putting in place an unprecedented stream of intelligence.<|endoftext|>"That policy strengths are important strategically. our partnership with the Center for Health Transformation will provide answers to these questions and our support will make the system work as a starting foundation for a market-driven national and market-based healthcare system. We've a world-famous health insurance plan for the millions who are currently lacking coverage," said state Senator Barbara W. Bush (due action) .<|endoftext|>Meanwhile I have spent my next four years women battling to turn in the ideal world into my daughter's .<|endoftext|>Foot-<|endoftext|>
\end{tcblisting}

\textbf{Di-LADD 32}
\medbreak

\begin{tcblisting}{genbox}
<|endoftext|> was ruled out about a while McCoy returned to the hospital Tuesday after a brief simulated game after sitting out four straight knockouts.<|endoftext|>Member of Creve told FOX: " Obviously, it's a surreal memorandum we look forward to signing.<|endoftext|>Willem Daughtry star in the production, which opens nationwide for the fifth season at the theater.<|endoftext|>"So there's better in the bedroom?"<|endoftext|>Shahid is a former commander in Afghanistan and Ahmad injaid is a local cleric.<|endoftext|>Under European football's Fifa Participation, the lionesses of the World Cup, both matnis, will be accused of being a sor<|endoftext|>
\end{tcblisting}

\paragraph{Generated sequences on OWT, uncurated.}

\textbf{MDLM}
\medbreak

\begin{tcblisting}{genbox}
's an open horizontal opening of the median opening to prevent bee emphyozence, meaning the fissured or rhizome makes for more honey."

In fact, giant pandas are far more squishping than their fishers are known to be. In theory, the possible case for the kind of mystique is full of holes: Originally, giant pandas were never made out of silk, and Chinese silk handles without a handle.

Alison Aleksand, a zoologist at the Smithsonian's National China Center in The Smithsonian, and two long colleagues think giant pandas may a developed a "nonpartite" capacity for concentrates."It's not as great with smaller doses as large rice pottery, speeds we've never reached," Ahdose says. "I'm here to preserve that complexity in the book."Since the Iraq Gulf war and its aftermath, I have frequently noticed that Israeli politics has increasingly disastrously preferred Israel, especially Iran for its ideological partners. The Republican Party is so sure on what they think of Israel that long-serving advisors, such as Robert Hagan and Colin Powell, wouldn't hurt the party by the president's actions.

Unlike Iran policy, though, yet it seems that Netanyahu's his boss. For me, last week Netanyahu's job search (in a sense) best suited Israeli Benjamin Netanyahu. This made perfect sense given Netanyahu's mutual deference to some of Israel's friends and even the president involved with. But because of the 1995 Jerusalem accords, on the other hand, it's down to who Israel decides to rest on. With Iran stuck, and depending upon U.S. and Israeli policy there, to lay would make a choice is difficult.

The same goes for Netanyahu, but especially for U.S. policy there. Iran is probably fallen into a version of this. That's so because during the height of the 2006 invasion of Iraq, America's strategists feared that Obama's Panetta would be used as a tool of absolution from accountability for his shortcomings there. He intended to memorandize Iraq by 2008, perhaps making Iran go over it with some personal concessions. But that seems to have been canceled even if he faced a vigorous attack from someone more likely to keep his adept mouth shut blowing his nose.

And so to insist on warping Iran for geopolitics was against him. When his country was wildly hawk, Netanyahu had precious few choices. One cannot truly say what military he would not have had made the lengths to which he would pick Israel
\end{tcblisting}

\textbf{Co-LADD 16}
\medbreak

\begin{tcblisting}{genbox}
for the House Science and Technology Committee, which now falls off the Senate oversight panel, on safeguard of church and speech rights.

Advertisement Continue reading the main story

On the mysterious logins of phone calls at the White House in April 2003, top staff officials said, "To state the very simple fact: over a quarter are being harmed, including personal data, in a few minutes, on average, and that cuts off tens of thousands of our staff at the White House."

Photo

A veteran questioned whether some of the administration's surveillance practices were a "political first," though called members of Congress "misguided" and even "stupid."

Yet said the soldier, who has always cautioned against adventurism, suggesting the government's only way to gain more damage control over American lives than to "do domestic terrorism."

"You've got computerized phone records and documents and business records," he said. "This isn't an ordinary thing you can claim to have perfected."

The question of what's reported when people contact terrorism, which has been the center of intense controversy, could face major repercussions recently.

Take the Justice Department's scandal when House Republicans went for Binney in February over astonishing information about the administration's "potential misbehavior" on issues of religion and political affiliations.

House Republicans criticized the heads of dozens of member associations, although Democrats said they were the correct way: If government surveillance sent them upset.

"You want our nation the only 24-hour line-and to get to the limit, that's right, to respect one of America's most important freedoms-than is displacing one over another," the Democrats said.

What do you think that at a higher level, even more sophisticated cyber spies could have reacted by asking: What's an Acton Edwards opinion that caused all this in the first place?

Senator Grayson Grayson of Tampa, Fla., a vocal conservative Republican, accused Turner of a conspiracy on Facebook to "bring down" Osama bin Laden and spread other things.

In 2004, Turner's quick, if somewhat inaccurate, correction on, the Pentagon's upcoming military budget plan that was aired at CNN amounted to a note Turner has since deemed "unworthy."

Treyer had spent the best parts of his military career online exploring the Army and reading his own book but he noticed the constant plague on American aides and false flags.

In his remarkable biography, he detailed how many of the nation's former top commanders were loved or revered because they took great pride in their service.

"That doesn
\end{tcblisting}

\textbf{Co-LADD 64}
\medbreak

\begin{tcblisting}{genbox}
The changesal cars also are homes that provide passengers to gain more safety and access, while offering a number of courses that provide furtherance. The car driver can also help ease the day-to-day dealing with tasks that aren't important to the homeowner, and allows them to choose how to use the car to support their job.

People Who Take Traditional Home Ownership. In large part, many of their homes assume traditional ownership, and if buyers take taxes and loan to a lower or lower debt-extracting house, the majority of their homeownership will pay loan debt again. The IRS also allows users to store their income, whether it's an insurance or real estate loan. They may toss a dime into the homeowner's pocket for income, instead of pay back interest.

The Fed's attempts at promoting home ownership also become a factor. Many Fed-driven schemes change the way that consumer loans carry their money toward mortgage payments.

This may cause an eventual housing bubble, allowing more housing debt to trickle down to homeowners.

Real estate loans also help augment borrowing, increasing the household income stream, thereby adding the revenue available to finance various activities within the corporate and government sectors. While most non-owners required to write down mortgages have been homeowners, they have often taken decisions on an individualist basis. Simultaneously, does anyone imagine using their own assets to fund improvements, improving infrastructure, among other things?

Libertarian and mainstream government policy revel in shining how much upon the home and on the aggregate, ignoring that it is good for it. In fact, some conservative reformers believe that the government is encouraging homeowners to get its hands off their homes, because they are preventing them from getting into trouble. In fact, many libertarians believe-women should bother all the time with what is necessary, besides profit. This is the only way we have a fully competitive system from home to home, owing to the rigorous, sound, concerns about competition.

Libertarians would generally agree that government intrusion in the home industry as not so valuable to their interests, a result that it is to have government enter the mortgage business. Other triggers targeted by modern modern technology are swiping or checkout crawling practices, because homeowner's government enters their home when considering a mortgage. In fact, most homeowners have no mortgage policy because purchases are carried out in their home and then "off the books," until they obtain a new government license. In mainstream housing, frequently, the government accepts a loan from a group of homeowners that are serviced on the basis of loans in by the mortgage lender.
\end{tcblisting}

\textbf{Di-LADD 16}
\medbreak

\begin{tcblisting}{genbox}
the frame count."

So do you envision seeing the recent SteamWorks update when the Patrice 6 engine pops ups on PC?

I don't know if any company said that, but for VR, you have to be able to log into the My Touch section and keep working just fine-as it actually did.

"So we're VR studio, and we were very excited about that point. We therefore wanted to create a game that could take a fragment, and just have an extra sense of information like this, to be analogous to that information you see FIDEIV. We have some really extremely interesting games to create. Probably they will be truly high of pressure.

"A bit more interesting is how good Indiana Jones dataflow is like. At things like Unreal that it's going to be awesome. And I know there are many to that, but hopefully it's these tracking feedback that we've got internally about some of the ones, we're still lacking the chance that eventually we can have something that makes core players [other]."

Port pauses, and then answers this question.

"And finally, if you move CSGO what this would mean for FIDEIV would be interesting, so to me that's interesting. Valve certainly has interest in doing that. We haven't had to answer that yet, because there's so much to work to learn."

PollenAs a blocky concept model we discover instead another possibility for a new sporting gun, crafted from a semisynchronous chamber like a Glock 3. After a unique true story to an inventor with military abilities, we come to a surprisingly accurate story.

Gannon is undoubtedly someone else's true expertise, but there is nothing that we do know of any firearm. Unlike only real guns, they feed specially made accelerometers that are hooked pieces in concert to attach to each other. The Glock gun can reliably hit anyone in the world; in the past, its entire guns had limited shooting capacity in many locations don't have challenges in a physics of the real gun (Germany)?.

A variety article on this topic is available in the magazine Suzy Parademons: Fat Creamery or Foreign.Update: Building P38 in Paris was a coup for Islam by a member of the Christian Offates. They are notoriously conservative New band, loosely organized for more than 20 years, is the main enemy of Breitbart and Islam's political will, critics, and trading partners at Europe. Politically, many of these Germans now perform the worst acts on migrants, actually, let us say
\end{tcblisting}

\textbf{Di-LADD 64}
\medbreak

\begin{tcblisting}{genbox}
student shortages and the rise of Vietnam Veterans' Union troops.

"Good Bill Burns, or would you tell the friggin cowboy?" Burns replied, "DeChote. "

Two years later, Burns was a Second Amendment champion who "facehooting" in a National Guard uniform.

Photo Gallery

When police register at Barthol's in 2010 in Lovato, Arizona, the office opposite the words "Joe the"

 is on a small screen.

Come back in early, hear the owner welcome a copy of survival light novel Joe the Movie, the stand-in movie about a shooter who appears in a fictional way.

When Adam and Anthony were shot by the tavern, a local photographer said he saw the fate of slain Gander Elementary School in Staunton County.

Retired Lt. Mark George Ryanes rides a shotgun. "can he hide with any trident at all," said Oklahoma historian "John Webster."

The 60-year combat vet said when an Iraq War man committed suicide, he made him "advent mates" at the 1943 song "I Beloved" the episode of American Armed Forces Bistro.

(Editing by Alan RaeAP Conservative Annabel Woodman to scrap the charge for free petrol for cars in Birmingham has attracted opposition from a bid to slash local charges and bring out of his party's seat too, MPs said.

Around 24,000 people turned by in the last week, arguing the petrol charge would be too high for a car dealer in the constituency East.

Lib Dem spokesman Mark Twigg called for local authorities to "get back to paying support fees," and said Lewiston taxpayerite would pay the annual fees for petrol, but the charge was already well worth being used as a charge.

"She doesn't pay you a penny other than one year support of house and car and it would be an element in that problem within the government," he told ITV News, adding that Labour would also make a further fairer Tory. "It certainly is the link to something that has been lost in the past, when a tax for food and drinks was introduced."

Most people are voting Independently but have not yet known for Labour.

Getty Panel models predicting unemployment: Over the next 20 years, around 102,000 will

She insisted Birminghamians would be in the jobs. "Times are changing. Early studies say less is going to be in the equation. We're not seeing much less than the stats," she told ITV
\end{tcblisting}

\end{document}